\documentclass[runningheads]{llncs}
\usepackage{graphicx}

\usepackage{tikz}
\usepackage{comment}
\usepackage{amsmath,amssymb} 
\usepackage{color}
\usepackage[pagebackref,breaklinks,colorlinks]{hyperref}

\usepackage{algorithm}
\usepackage{algpseudocode}
\usepackage{mathtools}
\usepackage{booktabs}
\usepackage{multirow}
\usepackage{xspace}

\usepackage[capitalize]{cleveref}
\crefname{section}{Sec.}{Secs.}
\Crefname{section}{Section}{Sections}
\Crefname{table}{Table}{Tables}
\crefname{table}{Tab.}{Tabs.}

\makeatletter
\DeclareRobustCommand\onedot{\futurelet\@let@token\@onedot}
\def\@onedot{\ifx\@let@token.\else.\null\fi\xspace}
\def\etal{et al\onedot}

\DeclareMathOperator*{\argmax}{arg\,max}
\DeclareMathOperator*{\argmin}{arg\,min}
\DeclarePairedDelimiter\floor{\lfloor}{\rfloor}
\DeclarePairedDelimiter\ceil{\lceil}{\rceil}

\newcommand{\norm}[1]{\left\lVert#1\right\rVert}
\algnewcommand{\LineComment}[1]{\State \(\triangleright\) #1}

\makeatletter
\newcommand{\algmargin}{\the\ALG@thistlm}
\makeatother
\algnewcommand{\parState}[1]{\State%
    \parbox[t]{\dimexpr\linewidth-\algmargin}{\strut\hangindent=\algorithmicindent \hangafter=1 #1\strut}}

\usepackage[accsupp]{axessibility}  

\begin{document}
\pagestyle{headings}
\mainmatter
\def\ECCVSubNumber{6269}  

\title{Soft Masking for Cost-Constrained Channel Pruning} 

\titlerunning{Soft Masking for Cost-Constrained Channel Pruning}
%
\author{Ryan Humble\inst{1}\thanks{Work performed during a NVIDIA internship} \and
Maying Shen\inst{2} \and
Jorge Albericio Latorre\inst{2} \and Eric Darve\inst{1} \and Jose Alvarez\inst{2}}
\authorrunning{R. Humble et al.}
%
\institute{Stanford University, Stanford CA 94305, USA \\
\email{\{ryhumble,darve\}@stanford.edu} \and
NVIDIA, Santa Clara CA 95051, USA \\
\email{\{mshen,jalbericiola,josea\}@nvidia.com}}
\maketitle

\begin{abstract}
Structured channel pruning has been shown to significantly accelerate inference time for convolution neural networks (CNNs) on modern hardware, with a relatively minor loss of network accuracy. Recent works permanently zero these channels during training, which we observe to significantly hamper final accuracy, particularly as the fraction of the network being pruned increases. We propose Soft Masking for cost-constrained Channel Pruning (SMCP) to allow pruned channels to adaptively return to the network while simultaneously pruning towards a target cost constraint. By adding a soft mask re-parameterization of the weights and channel pruning from the perspective of removing input channels, we allow gradient updates to previously pruned channels and the opportunity for the channels to later return to the network. We then formulate input channel pruning as a global resource allocation problem. Our method outperforms prior works on both the ImageNet classification and PASCAL VOC detection datasets.
\end{abstract}

\section{Introduction}
Deep neural networks have rapidly developed over the last decade and come to dominate many traditional algorithms in a wide range of tasks. In particular, convolutional neural networks (CNNs) have shown state-of-the-art results on a range of computer vision tasks, including classification, detection, and segmentation. However, modern CNNs have grown in size, computation, energy requirement, and prediction latency, as researchers push for accuracy improvements. Unfortunately, these models can now easily exceed the capabilities of many edge computing devices and requirements of real-time inference tasks, such as those found in autonomous vehicle applications.

\begin{figure}[t]
  \begin{center}
    \includegraphics[width=\columnwidth]{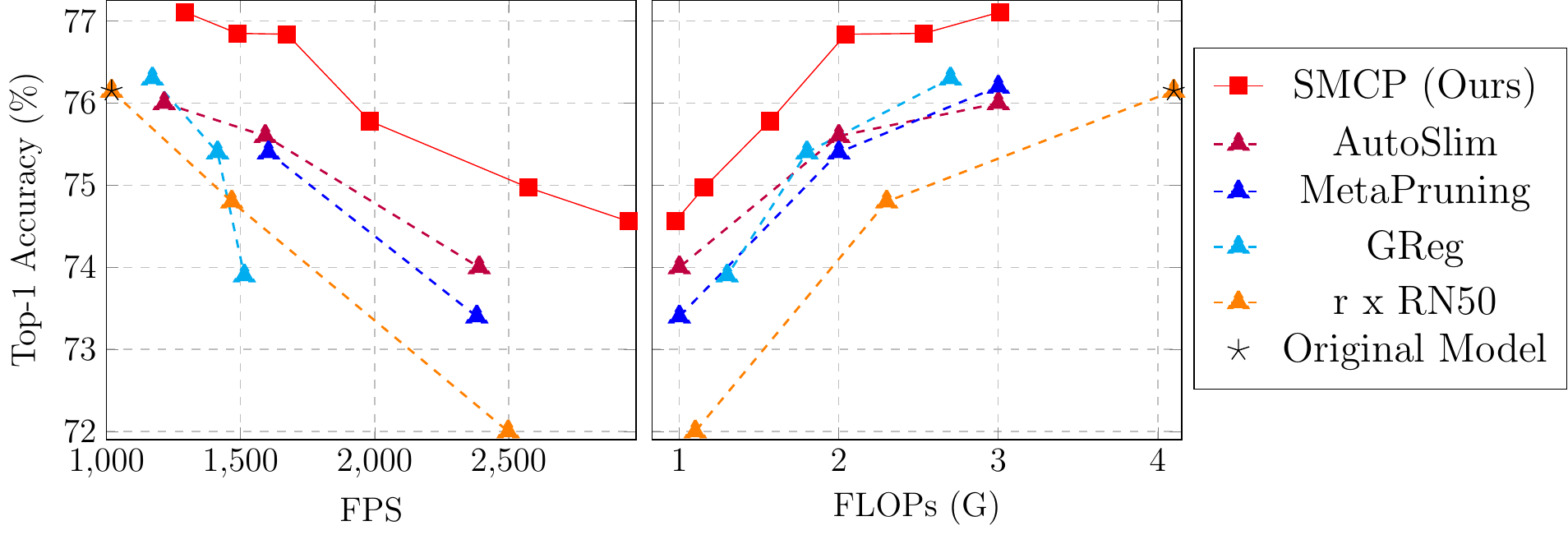}
  \end{center}
  \caption{Top-1 accuracy tradeoff curve for pruning ResNet50 on the ImageNet classification dataset using a latency cost constraint. Baseline is from PyTorch~\cite{paszke_2019} model hub. Accuracy against FPS speed (left) and FLOPs (right) show the benefit of our method, particularly at high pruning ratios. For FPS, top-right is better. For FLOPs, top-left is better. FPS measured on an NVIDIA TITAN V GPU.}\label{fig:resnet50_baseline_compare} 
\end{figure}

Since neural networks have been shown to be heavily over-parameterizered \cite{zhang_2017}, one popular method for reducing the computation and prediction latency is to prune (or remove) portions of the neural network, ultimately yielding a model with fewer parameters. Due to the strict requirements for many deployment applications, a large fraction of the parameters often must be removed; we focus on this regime, which we refer to as the high pruning ratio regime. Towards this aim, many pruning methods have been proposed to identify and remove those parameters that are least important for inference~\cite{alvarez_2016,he_2017_pruning,luo_2017,lecun_1989,molchanov_2019,ye_2018,yu_2018}. Since each layer of the network involves a different computation and associated computational burden, each parameter does not contribute equally to the final network inference cost, typically measured as FLOPs or latency, so more recent works have focused on pruning the network subject to explicit cost constraints. To maximize inference speedup on modern hardware (e.g., GPUs), these works largely focus on channel pruning~\cite{li_2020,luo_2017,molchanov_2019,shen_2021,wang_2021,yu_2019}.

However, in general, existing pruning works permanently remove the network parameters along these channels, zeroing the network weights and preventing the channel from being used during the rest of training. Particularly at high pruning ratios, where a significant fraction of the total channels in the network must be removed, the decisions on which channels to remove early during pruning are potentially myopic. Moreover, as a large number of channels are removed, the gradients to the remaining channels in each layer are significantly disrupted and can grow quite substantially due to the batch normalization layers ubiquitous in modern CNNs. This interferes with both network training and the identification of which further channels to remove.







In this work, we introduce a novel channel pruning approach for neural networks that is particularly suitable for large pruning ratios. The core of our approach relies on regularly rewiring the network sparsity, through soft masking of the network weights, to minimize the accuracy drop for large pruning ratios. The introduction of soft masking allows previously pruned channels to later be restored to the network, instead of being permanently pruned. Additionally, to mitigate the effect of large gradient magnitudes caused by removing many channels, we incorporate a new batch normalization scaling approach. Lastly, we formulate channel pruning under a cost constraint as a resource allocation problem and show it can be efficiently solved. All together, we refer to this method as Soft Masking for cost-constrained Channel Pruning (SMCP).

Our main contributions are:
\begin{enumerate}
    \item We demonstrate that a network's channel sparsity can be adaptively rewired, using a soft mask re-parameterization of the network weights, and that this requires channel pruning to be performed along input, instead of output, channels, see \cref{sec:soft_mask}.
    \item We propose a new scaling technique for the batch normalization weights to mitigate a gradient instability at high channel pruning ratios, see \cref{sec:bn_scaling}.
    \item We perform channel pruning subject to a cost constraint by encoding it as a resource allocation problem, which automatically allocates cost across the network instead of relying on manual or heuristic-based layer-wise pruning ratios. We show this allocation problem is a variant of the classic 0-1 knapsack problem, called the multiple-choice knapsack problem~\cite{sinha_1979}, which can be efficiently solved for our experiments, see \cref{sec:cost_constrained_pruning}.
    \item We analyze our method's accuracy and cost improvements for the ImageNet and PASCAL VOC datasets for ResNet, MobileNet, and SSD architectures. We outperform prior pruning approaches, as shown in~\cref{fig:resnet50_baseline_compare} and more extensively in~\cref{sec:results}. In particular, at high pruning ratios for ResNet50/ResNet101 on ImageNet, SMCP can achieve up to an additional \( 20\% \) speedup at the same Top-1 accuracy level or up to a \( 0.6\% \) Top-1 accuracy improvement at the same FPS (frames per second). SMCP can also prune an SSD512 with a ResNet50 backbone to achieve a speedup of \( 2.12 \times \), exceeding the FPS (frames per second) of the smaller SSD300-ResNet50 model by \( 12 \% \), while simultaneously improving on the mAP of the baseline model.
\end{enumerate}


\section{Related Work}
\subsection{Soft pruning}
Most pruning methods start with a dense pretrained network and prune iteratively over a schedule to obtain a final network with the desired cost, where at each pruning step parameters are permanently zeroed (or masked). This effectively limits the model capacity as pruning occurs. Stosic and Stosic~\cite{stosic_2021} argue that preserving the larger model capacity is critical to sparse model training by forming new paths for optimization that are not available for permanently pruned networks; they suggest it is important to allow gradient flow to previously pruned parameters and to rewire the sparsity occasionally.

Along these lines, several works have proposed soft pruning methods where parameters can be pruned and later unpruned if desirable. He \etal~\cite{he_2018} zero weights during pruning but allows gradients to update them in an effort to maintain model capacity.
Dettmers and Zettlemoyer~\cite{dettmers_2019}, Evci \etal~\cite{evci_2020}, Mostafa and Wang~\cite{mostafa_2019}, and Wortsman \etal~\cite{wortsman_2019} allow previously pruned weights to be regrown. Kusupati \etal~\cite{kusupati_2020} used a soft thresholding operator to achieve state-of-the-art results for unstructured and low-rank structured pruning.
Kang and Han~\cite{kang_2020} introduces soft channel pruning by adding a differentiable mask in the batch normalization layers; however, their approach is limited to an implicit cost constraint on the total number of neurons. Our approach though is most similar to Guo \etal~\cite{guo_16}, Lin \etal~\cite{lin_2020}, De Jorge \etal~\cite{dejorge_2021}, and Zhou \etal~\cite{zhou_2021}, which explicitly or implicitly use the Straight-through Estimator (STE)~\cite{bengio_2013} to adaptively prune parameters during training. The first three target unstructured sparsity, and the last targets N:M structured sparsity. 
In our work, we extend the use of the STE to channel pruning, show this requires pruning to be formulated along input channels, and embed this soft masking into a general-purpose, explicit cost-constrained formulation.

\subsection{Cost-constrained and structured pruning}
The goal of most pruning methods is to maximize network accuracy subject to low memory, computation, and/or latency requirements. Although unstructured sparsity approaches have proven to be very successfully in removing upwards of 95\% of weights without affecting network accuracy~\cite{han_2015}, modern hardware has poor support for unstructured sparsity and therefore this rarely translates to actual speedup. Therefore, it is common to choose a pruning sparsity structure that can actually be accelerated in hardware, typically channel pruning for CNNs. There is now some hardware support for other sparsity structures, such as the N:M structured sparsity of~\cite{mishra_2021_ampere}, but we limit our focus to channel pruning in this work. Both Li \etal~\cite{li_2020} and Yang \etal~\cite{yang_2018,yang_2021} select the best constraint-abiding network from a large number of candidate networks, which can be prohibitively expensive. Yu and Huang \etal~\cite{yu_2019}, Tan \etal~\cite{tan_2019}, and Wu \etal~\cite{wu_2020} pose cost-constrained optimization problems but use a greedy selection or cost-aware importance score to approximately select the best channels to prune. Chen \etal~\cite{chen_2018} presents a Bayesian optimization approach to determine compression hyperparameters that satisfy a cost constraint while maximizing network accuracy. Liu \etal~\cite{liu_2019} linked network pruning to Neural Architecture Search (NAS), arguing that the resulting pruned architectures are the novel contribution instead of the trained weights themselves. However, most NAS methods, such as those in~\cite{dai_2019,dong_2018,su_2021}, remain more computationally expensive than network pruning approaches. Our approach is most similar to the concurrent work of Shen \etal~\cite{shen_2021}, called HALP, which also poses a cost-constrained resource allocation problem. There are however several major differences. First, we reduce our allocation problem to the multiple choice knapsack problem~\cite{sinha_1979} and solve it with a meet-in-the-middle algorithm, which provides both optimality guarantees and efficient (\(<1\) second) solutions for general cost-constraints. HALP solves their allocation problem with a custom augmented knapsack solver, which gives no optimality guarantees and requires significant extra computation (\( 1+ \) minute for each pruning step on ResNet50~\cite{he_2016}, even after a large GPU-specific neuron grouping step). Second, our method uses soft input channel masking as opposed to the permanent output channel pruning of HALP; we show this change alone yields performance gains in~\cref{sec:ablation_study}. Lastly, we use a new batch normalization scaling technique to stabilize training at high pruning ratios.




\subsection{Pruning impact on batch normalization layers}
Channel pruning can have a significant impact on the batch normalization statistics, which therefore strongly affects the network gradients to the remaining channels. This effect is particularly pronounced at high pruning ratios, since a large number of channels are being removed from most layers. Several pruning methods note this phenomenon and describe mitigation strategies. Li \etal~\cite{li_2020} demonstrated the need to update the batch normalization statistics after pruning, as they can be significantly impacted, before evaluating possible pruned candidate networks. This approach does not however alleviate the issue of large gradients. Instead of immediately removing pruned weights and incurring the disruption, Wang \etal~\cite{wang_2021} slowly regularized them away, noticing significant performance gains particularly at high pruning ratios. They do not connect this to a sudden change in batch normalization statistics and gradients caused by pruning. They also use a non-gradient based importance so the impact on the importance of the remaining parameters is somewhat subdued. Since we are adaptively adjusting the sparsity and want to preserve the ability for pruned weights to become later unpruned, we do not want to regularize away pruned weights. We instead adopt a scaling technique on the batch normalization weights to stabilize training at high pruning ratios.

\subsection{Parameter importance scoring}
In order to decide which parameters of the network can be pruned while least harming network accuracy, most pruning methods define an importance for each parameter (or set of parameters) that approximates the effect of removal on the network's loss. Many importance scores have been proposed, largely falling into three groups: (i) based on weight magnitude~\cite{alvarez_2016,han_2016,li_2017,liu_2017,ye_2018,yu_2018}; (ii) based on a reconstruction-based objective~\cite{he_2017_pruning,luo_2017}; and (iii) based on network gradients~\cite{lecun_1989,molchanov_2019,molchanov_2017}. We adopt the Taylor first-order importance~\cite{molchanov_2019} due to its computational simplicity and its strong correlation with the true impact on the network's loss.

\section{Soft masking for cost-constrained channel pruning}
We propose a novel input channel pruning approach targeted towards high pruning ratios. Our method is initialized with a pretrained CNN model, and the desired network cost function and target cost constraint. We first re-parameterize the network weights with input channel masking variables, as shown in~\cref{sec:soft_mask}, to enable adaptive channel pruning. Then, after a warmup period, we iteratively prune every \( r \) minibatches by solving a resource allocation optimization problem, discussed in~\cref{sec:cost_constrained_pruning}, to update the channel masks. After each mask update, we apply the batch normalization scaling described in~\cref{sec:bn_scaling}, which stabilizes training at high pruning ratios. Finally, we fix the masks for a cooldown and fine-tuning period. We present the full algorithm and pseudocode in~\cref{sec:smcp}.


\subsection{Soft input channel pruning}\label{sec:soft_mask}
\begin{figure}[t]
  \begin{center}
    \includegraphics[width=0.30\columnwidth]{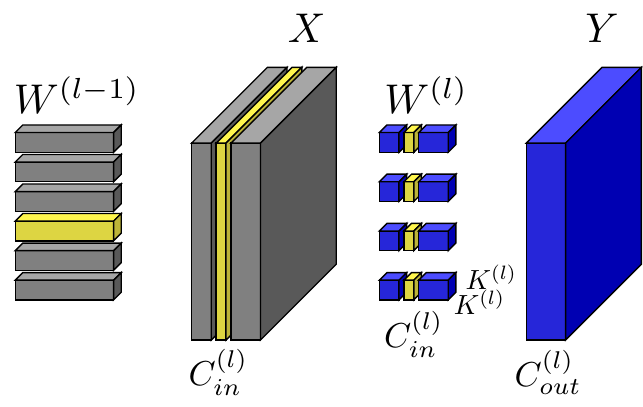}
  \end{center}
  \caption{Input channel pruning of a convolutional layer. Removing an input channel from weight \( W^{(l)} \in \mathbb{R}^{C^{(l)}_{out} \times C^{(l)}_{in} \times K^{(l)} \times K^{(l)}} \) in layer \( l \) removes the corresponding channel in the input feature map \( X \) and the corresponding output channel in the previous weight \( W^{(l-1)} \). The shape of the output feature map \( Y \) is unaffected.}
  \label{fig:input_pruning}
\end{figure}

We specifically consider input channel pruning, as previously done in~\cite{he_2017_pruning} and shown in~\cref{fig:input_pruning}, where we mask and later remove input channels to sparsify the CNN. As we will shortly show, channel pruning with a soft mask re-parameterization requires it to be done along input channels, as this approach does not work when performing output channel pruning. This is a departure from the many output channel pruning approaches. From a global view of network sparsity, pruning one layer's input channel is equivalent to pruning the previous layer's output channel; however, the approaches are distinct when considering the effect on each individual layer.

For soft input channel masking, we consider a neural network with weights \( W = \{ W^{(l)} \} \), where \( W^{(l)} \in \mathbb{R}^{C^{(l)}_{out} \times C^{(l)}_{in} \times K^{(l)} \times K^{(l)}} \) is the weight for layer \( l \) of the network and has \( C^{(l)}_{in} \) input channels and \( C^{(l)}_{out} \) output channels. To allow input channels to be pruned and later unpruned, we introduce an input channel mask \( m^{(l)} \in \{0,1\}^{C^{(l)}_{in}} \) for each layer \( l \). Using these masks, we re-parameterize the weights so that the network's sparse weights are
\begin{equation}\label{eqn:mask_reparam}
    \widetilde{W}^{(l)} = W^{(l)} \odot m^{(l)}.
\end{equation}
where \( m^{(l)} \) is broadcasted to match the shape of \( W^{(l)} \). Instead of permanently zeroing a channel when pruning, the underlying network weights can be preserved and merely the masks set to zero. This has two distinct advantages. First, it helps preserve the full capacity of the original model while training towards a sparse model. Second, by allowing channels to be restored to their original values at a later time, poor early decisions on where to allocate the sparsity across the layers can be undone. This is particularly important for high pruning ratios where a large portion of the network's channels must be removed.

As written though, our masking definition would define the gradient with respect to \( W^{(l)} \) as \( g_{W^{(l)}} = g_{\widetilde{W}^{(l)}} \odot m^{(l)} \). This masks the gradients as they flow back to the completely dense weights \( W \), rendering masked weights unused in the forward pass and left untouched by the backward pass. Following the argument by Stosic and Stosic~\cite{stosic_2021} that updating parameters not currently participating in the forward pass offers additional optimization paths that improve training of sparse networks, we adopt the Straight-through Estimator (STE)~\cite{bengio_2013}. The STE has been successfully used in model quantization~\cite{rastegari_2016} and Ampere 2:4 structured pruning~\cite{zhou_2021} for sparse parameter updates. The STE defines the gradient as
\begin{equation}\label{eqn:ste}
    g_{W^{(l)}} = g_{\widetilde{W}^{(l)}},
\end{equation}
where gradients on the sparse weights pass straight through to the underlying, dense weights. Note that we still use the masks when computing the gradient with respect to the input feature map of the layer.

However, for this STE to have a useful impact in a modern CNN with the ubiquitious Conv-BN-ReLU pattern, it requires that channel pruning must be posed as input-oriented. Since \( g_{\widetilde{W}^{(l)}} \) is defined by a matrix multiplication using the input feature map and the gradient of the output feature map, a masked input channel still receives non-zero gradients, except under a few edge cases. If we had instead masked output channels, the elements of \( g_{W^{(l)}} \) would be either \( 0 \) or \( \infty \), depending on the value of the batch normalization bias. Alternatively, if we instead tried to directly mask the batch normalization weight \( \gamma^{(l)} \) and bias \( \beta^{(l)} \) to emulate pruning the channel, we would get \( g_{\gamma^{(l)}} = g_{\beta^{(l)}} = 0 \) due to the ReLU. In either of these cases, the gradient \( g_{W^{(l)}} \) is not useful.

Finally then, for input channel pruning with soft masking, we define the importance of each input channel, a proxy for the effect of removing this channel on the network's loss, according to the group first-order Taylor importance of~\cite{molchanov_2019}:
\begin{equation}\label{eqn:taylorfo_channel}
    \mathcal{I}^{(l)}_i = \left| \sum_{o,r,s} W^{(l)}_{o,i,r,s} g_{W^{(l)}_{o,i,r,s}} \right|
\end{equation}
where \( \mathcal{I}^{(l)}_i \) is the importance of the \( i \)th input channel to layer \( l \). Under certain conditions, this is in fact equivalent to the first-order batch normalization-based Taylor importance of~\cite{molchanov_2019}, as shown in the supplementary materials.

\subsection{Batch normalization scaling\label{sec:bn_scaling}}
When channel pruning at high ratios, there are many layers where a significant number of channels must be pruned. As a result of pruning these channels, either by zeroing them out or by applying masking, the subsequent gradient magnitudes to the remaining unpruned channels can be excessively large, which we show in the supplementary materials. We propose a batch normalization scaling technique that adjusts the batch normalization weight \( \gamma^{(l)} \) of layer \( l \) to mitigate large gradients and stabilize the network sparsity and training. Specifically, we scale \( \gamma^{(l)} \) according to the fraction of channels left unpruned by the current input channel mask \( m^{(l)} \in \{0,1\}^{C^{(l)}_{in}} \)
\begin{equation}\label{eqn:bn_scaling}
    \gamma^{(l)} \gets \gamma^{(l)}_{\textit{orig}} \frac{\sum_i m^{(l)}_i}{C^{(l)}_{in}}.
\end{equation}
In practice, we always treat \( \gamma^{(l)}_{\textit{orig}} \) as the parameter under optimization and vary a scaling variable \( s^{(l)} \) to adjust the weight used by the network.

Moderating gradient magnitudes is particularly consequential since we employ the gradient-based importance score shown in~\cref{eqn:taylorfo_channel}. Even without soft masking and the STE, the large gradients cause importance accumulation in the remaining channels as pruning iteratively proceeds, artificially inhibiting additional channels in the layer from being pruned. When employing soft masking without this scaling technique, the large gradients cause large network sparsity thrashing. For example, if at one pruning iteration a large number of the channels are pruned, the importance to every channel, not only those left unpruned, is boosted by the resulting large gradient magnitudes. At the very next pruning iteration, those channels appear quite important and are restored to the network, causing other portions of the network to be pruned to still meet the cost constraint. This can oscillate, inhibiting network convergence and the final network accuracy. Moreover, for architectures in which pruning entire layers is possible, such as ResNet due to the skip connections, the infinite gradient magnitudes cause numerical overflow in updating the weights or even calculating the importance of channels. As shown in our experiments in~\cref{sec:results}, the proposed batch normalization scaling is crucial to overcome these training issues.


\subsection{Cost-constrained channel pruning}\label{sec:cost_constrained_pruning}
At each pruning iteration, we seek to both minimize the impact on the network's loss as a result of pruning and sparsify the network towards the final cost constraint (e.g., latency constraints). We therefore formulate pruning as a cost-constrained importance maximization problem
\begin{align}\label{eqn:original_opt_problem}
    \max_{m^{(2)}, \dots, m^{(L)}} \quad & \sum_{l=1}^L \sum_{i=1}^{C^{(l)}_{in}} \mathcal{I}^{(l)}_i m^{(l)}_i \\
    \textrm{s.t.} \quad & \sum_{l=1}^L \mathcal{T}^{(l)}\left( \norm{m^{(l)}}_1, \norm{m^{(l+1)}}_1 \right) \leq \tau \nonumber \\
    & \norm{m^{(l)}}_1 \in \mathcal{P}^{(l)}, \nonumber
\end{align}
where \( L \) is the number of layers in the network, layer \( l \) has \( C^{(l)}_{in} \) input channels, \( \mathcal{I}^{(l)}_i \) is the importance of input channel \( i \) of layer \( l \), \( m^{(l)} \in \{0,1\}^{C^{(l)}_{in}} \) is the input channel mask for layer \( l \), \( \mathcal{T}^{(l)} \) is the cost function for layer \( l \), \( \tau \) is the cost constraint, and \( \mathcal{P}^{(l)} \) is the set of permitted values for the number of channels kept by mask \( m^{(l)} \). By definition, \( m^{(1)}_i = 1 \) and \( m^{(L+1)}_i = 1 \) since those are the unprunable inputs and outputs of the network. A complete derivation of~\cref{eqn:original_opt_problem} can be found in the supplementary materials, as well as a discussion on how to handle skip connections in architectures like ResNet~\cite{he_2016}.

The final constraint, on the set of permitted values \( \mathcal{P}^{(l)} \), is optional but useful in several situations. First, it can be used to disallow pruning the entire layer: by omitting \( 0 \) from \( \mathcal{P}^{(l)} \) we prevent $ m^{(l)} = 0$. As explained in the supplementary materials, layer pruning violates a key assumption of the derivation of~\cref{eqn:original_opt_problem}. Second, it can be used to ensure the number of remaining channels is hardware-friendly, such as \( 8 \times \) multiples for GPU tensorcores~\cite{nvidia_conv_userguide} with \( \mathcal{P}^{(l)} = \{0, 8, 16, \dots, \floor{C^{(l)}_{in}/8} \} \).

We can further reduce this to an optimization over only the number of channels \( p^{(l)} \), as the most important channels will always be kept in each layer:
\begin{align}\label{eqn:pruning_knapack_problem}
    \max_{p^{(2)}, \dots, p^{(L)}} \quad & \sum_{l=1}^L \sum_{i=1}^{p^{(l)}} \mathcal{I}^{(l)}_{(i)} \\
    \textrm{s.t.} \quad & \sum_{l=1}^L \mathcal{T}^{(l)}\left( p^{(l)}, \overline{p^{(l+1)}} \right) \leq \tau \nonumber \\
    & p^{(l)} \in \mathcal{P}^{(l)} \nonumber
\end{align}
where \( p^{(l)} = \norm{m^{(l)}}_1 \) and \( \mathcal{I}^{(l)}_{(i)} \) is the \( i \)th largest value in \( \mathcal{I}^{(l)} \). We also approximated the constraint using the current channel counts \( \overline{p^{(l)}} \) to decouple the cost impact of masks in consecutive layers, which is required to pose this as an example of the following class of optimization problems.

\subsubsection{Multiple-choice knapsack problem}\label{sec:multple_choice_knapsack}
The optimization problem in~\cref{eqn:pruning_knapack_problem} is an example of a generalization of the classic 0-1 knapsack problem called the multiple-choice knapsack (MCK) problem~\cite{sinha_1979}. We show this connection explicitly in the supplementary materials. The MCK problem takes the form
\begin{align}\label{eqn:multple_choice_knapsack}
    \max_{x} \quad & \sum_{l=1}^L \sum_{i=1}^{n_l} v_{l,i} x_{l,i} \\
    \textrm{s.t.} \quad & \sum_{l=1}^L \sum_{i=1}^{n_l} c_{l,i} x_{l,i} \leq C \nonumber \\
    & x_{l,i} \in \{0, 1\}, \quad \sum_{i=1}^{n_l} x_{l,i} = 1 \nonumber
\end{align}
where \( L \) is the number of groups, group \( l \) has size \( n_l \), and the items have value \( v_{l,i} \) and cost \( c_{l,i} \geq 0 \). The additional constraint relative to the classic 0-1 knapsack problem enforces that we select exactly one item from each group.

We solve \cref{eqn:multple_choice_knapsack} with a GPU-implemented meet-in-the-middle algorithm, presented in full in the supplementary materials. Our approach generalizes the standard meet-in-the-middle algorithm for the classic 0-1 knapsack problem, does not require integer costs, and very efficiently solves the MCK problem for our use cases. For example, for a ResNet50~\cite{he_2016}, our approach solves the MCK problem in under \( 1 \) second. We present more complete timing details in the supplementary materials.

\subsection{Overall method}\label{sec:smcp}
We present our full method in~\cref{alg:smcp}. We start with a pretrained network, layer-wise cost functions \( \mathcal{T}^{(l)} \), and a global cost constraint \( \tau \). We define our pruning schedule by: (i) \( K_w \): the number of warmup epochs before starting pruning; (ii) \( K_t \): the number of epochs after the warmup to reach the target cost \( \tau \); (iii) \( r \): the number of steps between recomputing the channel masks; and (iv) \( K_c \): the number of cooldown epochs where the masks are kept fixed. During those \( K_t \) epochs to reach the target cost, we define intermediate cost constraints \( \{ \tau_e \} \) using the exponential scheduler of~\cite{dejorge_2021}. Additionally, to stabilize the importance scores, which can be noisy due to the stochastic minibatches, we calculate and accumulate the importance score in~\cref{eqn:taylorfo_channel} every minibatch between pruning iterations according to the exponential momentum approach of~\cite{molchanov_2019}.

\begin{algorithm}[t]
    \caption{Soft masking for cost-constrained channel pruning}\label{alg:smcp}
    \textbf{Inputs}: Pretrained network weights \( W \), training set \( \mathcal{D} \), total number of epochs \( E \), pruning schedule \( (r, K_w, K_t, K_c) \), target cost \( \tau \)
    \begin{algorithmic}[1]
        \State Initialize masks \( m^{(l)} = 1 \)
        \State Re-parameterize the weights (\cref{eqn:mask_reparam})
        \State Train the network as usual for \( K_w \) epochs
        \State Calculate the pruning schedule \( \{\tau_e\} \)
        \For{epoch \( e \in [K_w, E - K_c) \)}
            \For{step \( s \) in epoch \( e \)}
                \State Perform the forward pass and backward pass, using \cref{eqn:mask_reparam,eqn:ste}
                \State Calculate and accumulate \( \mathcal{I}^{(l)}_i \) (\cref{eqn:taylorfo_channel})
                \If{\( s \% r = 0\)}
                    \State Solve the optimization problem (\cref{eqn:pruning_knapack_problem}) using target cost \( \tau_e \)
                    \State Update the masks \( m^{(l)} \) accordingly
                    \State Scale the BN weights \( \gamma^{(l)}\) (\cref{eqn:bn_scaling})
                    \State Reset the accumulated importance
                \EndIf
            \EndFor
        \EndFor
        \State Train the network as usual for \( K_c \) epochs
        \State Apply the masks to the weights permanently
        \State \Return Sparse network weights \( W \)
    \end{algorithmic}
\end{algorithm}

\section{Results}\label{sec:results}
We evaluate our method on both the ImageNet and PASCAL VOC benchmark datasets\footnote{Our code can be accessed at \url{https://github.com/NVlabs/SMCP}.}. Full details on training settings and architectures can be found in the supplementary materials. We use a latency cost constraint, defined by a layer-wise lookup table (LUT) as previously described in~\cite{yin_2020,shen_2021,yang_2018}. We target and measure latency speed on a NVIDIA TITAN V GPU with cudNN V7.6.5~\cite{chetlur_2014}.

\subsection{ImageNet results}\label{sec:imagenet_results}
We compare SMCP with several prior works on the ImageNet ILSVRC2012~\cite{russakovsky_2015} classification dataset.  In~\cref{tab:latency-aware-result-compare}, we compare the results of pruning ResNet50, ResNet101~\cite{he_2016}, and MobileNet-V1~\cite{howard_2017} at a number of pruning thresholds. We refer to SMCP-$X\%$ as retaining $X\%$ of the full model's original latency and calculate the frames per second (FPS) and speedup of the final network. For ResNet50, we show results for two different baseline models for a better comparison with prior works. The first baseline is from the PyTorch~\cite{paszke_2019} model hub, with a Top-1 accuracy of \( 76.15\% \); the second baseline is the one used as a baseline for EagleEye~\cite{li_2020} and has a Top-1 accuracy of \( 77.2\% \). We prune and fine-tune following the training setup of~\cite{nvidia_resnet_training_guide}.

\begin{table}[t]
    \caption{Pruning results on the ImageNet classification dataset considering two different ResNet50 baseline models as well as ResNet101 and MobileNetV1. We group results by those with similar FLOP counts, and refer to SMCP-$X\%$ as retaining $X\%$ of the full model's original latency. Results for prior works are as shown in~\cite{shen_2021}.}
    \label{tab:latency-aware-result-compare}
    \centering
    \begin{minipage}[t]{0.5\linewidth}
        \centering
        \resizebox{1.\columnwidth}{!}{
            \begin{tabular}{lccccc}
            \toprule
            \multirow{2}{*}{Method} & FLOPs & Top1 & Top5 &  FPS & \multirow{2}{*}{Speedup}\\
             & (G) & (\%) & (\%) & (im/s) &\\
            \midrule
            \midrule
            \multicolumn{6}{c}{\textbf{ResNet50}}\\
             No pruning & $4.1$ & $76.2$ & $92.87$ & $1019$ & $1\times$ \\
            \midrule
             ThiNet-$70$~\cite{luo_2017} & $2.9$ & $75.8$ & $90.67$ & - & - \\
             AutoSlim~\cite{yu_2019} & $3.0$ & $76.0$ & - &  $1215$ & $1.14\times$ \\
             MetaPruning~\cite{liu_2019} & $3.0$ & $76.2$ & - & - & - \\
             GReg-$1$~\cite{wang_2021} & $2.7$ & $76.3$ & - & $1171$ & $1.15\times$ \\
             HALP-$80\%$~\cite{shen_2021} & $3.1$ & $\mathbf{77.2}$ & $\mathbf{93.47}$ & $1256$ & $1.23\times$ \\
             \textbf{SMCP-$80\%$ (Ours)} & $3.0$ & $77.1$ & $93.43$ & $\mathbf{1292}$ & $\mathbf{1.27\times}$ \\
            \midrule
             $0.75 \times$ ResNet50~\cite{he_2016} & $2.3$ & $74.8$ & - & $1467$ & $1.44\times$ \\
             ThiNet-$50$~\cite{luo_2017} & $2.1$ & $74.7$ & $90.02$ & - & - \\
             AutoSlim~\cite{yu_2019} & $2.0$ & $75.6$ & - & $1592$ & $1.56\times$  \\
             MetaPruning~\cite{liu_2019} & $2.0$ & $75.4$ & - & $1604$ & $1.58\times$ \\
             GBN~\cite{you_2019_gate} & $2.4$ & $76.2$ & $92.83$ & - & - \\
             GReg-$2$~\cite{wang_2021} & $1.8$ & $75.4$ & - & $1414$ & $1.39\times$ \\
             HALP-$55\%$~\cite{shen_2021} & $2.0$ & $76.5$ & $93.05$ & $1630$ & $1.60\times$ \\
             \textbf{SMCP-$55\%$ (Ours)} & $2.0$ & $\mathbf{76.8}$ & $\mathbf{93.22}$ & $\mathbf{1673}$ & $\mathbf{1.64\times}$ \\
            \midrule
             $0.50 \times$ ResNet50~\cite{he_2016} & $1.1$ & $72.0$ & - & $2498$ & $2.45\times$ \\
             ThiNet-$30$~\cite{luo_2017} & $1.2$ & $72.1$ & $88.30$ & - & - \\
             AutoSlim~\cite{yu_2019} & $1.0$ & $74.0$ & - & $2390$ & $2.45\times$ \\
             MetaPruning~\cite{liu_2019} & $1.0$ & $73.4$ & - & $2381$ & $2.34\times$ \\
             GReg-$2$~\cite{wang_2021} & $1.3$ & $73.9$ & - & $1514$ & $1.49\times$ \\
             HALP-$30\%$~\cite{shen_2021} & $1.0$ & $74.3$ & $91.81$ & $2755$ & $2.70\times$ \\
             \textbf{SMCP-$30\%$ (Ours)} & $1.0$ & $\mathbf{74.6}$ & $\mathbf{92.00}$ & $\mathbf{2947}$ & $\mathbf{2.89\times}$ \\
            \midrule
            \midrule
            \multicolumn{6}{c}{\textbf{ResNet50 - EagleEye~\cite{li_2020} baseline}}\\
             No pruning & $4.1$ & $77.2$ & $93.70$ & $1019$ & $1\times$ \\
            \midrule
             EagleEye-3G~\cite{li_2020} & $3.0$ & $77.1$ & $93.37$ & $1165$ & $1.14\times$ \\
             HALP-$80\%$~\cite{shen_2021} & $3.0$ & $77.5$ & $93.60$ & $1203$ & $1.18\times$ \\
             \textbf{SMCP-$80\%$ (Ours)} & $3.1$ & $\mathbf{77.6}$ & $\mathbf{93.61}$ & $\mathbf{1263}$ & $\mathbf{1.23\times}$ \\
            \midrule
             EagleEye-2G~\cite{li_2020} & $2.1$ & $76.4$ & $92.89$ & $1471$ & $1.44\times$  \\
             HALP-$55\%$~\cite{shen_2021} & $2.1$ & $\mathbf{76.6}$ & $93.16$ & $1672$ & $\mathbf{1.64\times}$ \\
             \textbf{SMCP-$50\%$ (Ours)} & $1.9$ & $\mathbf{76.6}$ & $\mathbf{93.17}$ & $\mathbf{1706}$ & $\mathbf{1.67\times}$ \\
            \midrule
             EagleEye-1G~\cite{li_2020} & $1.0$ & $74.2$ & $91.77$ & $2429$ & $2.38\times$ \\
             HALP-$30\%$~\cite{shen_2021} & $1.2$ & $74.5$ & $91.87$ & $2597$ & $2.55\times$ \\
             \textbf{SMCP-$30\%$ (Ours)} & $1.1$ & $\mathbf{75.1}$ & $\mathbf{92.29}$ & $2589$ & $2.51\times$ \\
             \textbf{SMCP-$25\%$ (Ours)} & $0.9$ & $74.4$ & $91.98$ & $\mathbf{3102}$ & $\mathbf{3.01\times}$ \\
            \bottomrule
        \end{tabular}
        }
    \end{minipage}%
    \begin{minipage}[t]{0.5\linewidth}
        \centering
        \resizebox{.92\columnwidth}{!}{
            \begin{tabular}{lccccc}
                \toprule
                \multirow{2}{*}{Method} & FLOPs & Top1 &  FPS & \multirow{2}{*}{Speedup}\\
                 & (G) & (\%) & (im/s) &\\
                \midrule
                \midrule
                \multicolumn{5}{c}{\textbf{ResNet101}}\\
                 No pruning & $7.8$ & $77.4$ & $620$ & $1\times$ \\
                \midrule
                 Taylor-$75\%$~\cite{molchanov_2019} & $4.7$ & $77.4$ & $750$ & $1.21\times$\\
                 HALP-$60\%$~\cite{shen_2021} & $4.3$ & $\mathbf{78.3}$ & $847$ & $1.37\times$ \\
                 \textbf{SMCP-$60\%$ (Ours)} & $4.0$ & $78.1$ & $951$ & $1.53\times$ \\
                 HALP-$50\%$~\cite{shen_2021} & $\mathbf{3.6}$ & $77.8$ & $994$ & $1.60\times$ \\
                 \textbf{SMCP-$50\%$ (Ours)} & $\mathbf{3.6}$ & $77.8$ & $\mathbf{1016}$ & $\mathbf{1.64\times}$ \\
                \midrule  
                 Taylor-$55\%$~\cite{molchanov_2019} & $2.9$ & $76.0$ & $908$ & $1.47\times$ \\
                 HALP-$40\%$~\cite{shen_2021} & $2.7$ & $77.2$ & $1180$ & $1.90\times$ \\
                 \textbf{SMCP-$30\%$ (Ours)} & $2.6$ & $\mathbf{77.3}$ & $1273$ & $2.05\times$ \\
                 HALP-$30\%$~\cite{shen_2021} & $\mathbf{2.0}$ & $76.5$ & $1521$ & $2.45\times$ \\
                 \textbf{SMCP-$25\%$ (Ours)} & $\mathbf{2.0}$ & $76.8$ & $\mathbf{1535}$ & $\mathbf{2.48}\times$ \\
                \midrule
                \\
                \toprule
                \multirow{2}{*}{Method} & FLOPs & Top1 & FPS & \multirow{2}{*}{Speedup}\\
                 & (M) & (\%) & (im/s) &\\ 
                \midrule
                \midrule
                \multicolumn{5}{c}{\textbf{MobileNet-V1}}\\
                 No pruning & $569$ & $72.6$ & $3415$ & $1\times$ \\
                \midrule
                 $0.75 \times$ MobileNetV1 & $325$ & $68.4$ & $4678$ & $1.37\times$ \\
                 NetAdapt~\cite{yang_2018} & $284$ & $69.1$ & - & - \\
                 MetaPruning~\cite{liu_2019} & $316$ & $70.9$ & $4838$ & $1.42\times$ \\
                 EagleEye~\cite{li_2020} & $284$ & $70.9$ & $5020$ & $1.47\times$ \\
                 HALP-$60\%$~\cite{shen_2021} & $297$ & $\mathbf{71.3}$ & $5754$ & $1.68\times$ \\
                 \textbf{SMCP-$60\%$ (Ours)} & $356$ & $71.0$ & $\mathbf{5870}$ & $\mathbf{1.72\times}$ \\
                \midrule
                 MetaPruning~\cite{liu_2019} & $142$ & $66.1$ & $7050$ & $2.06\times$ \\
                 AutoSlim~\cite{yu_2019} & $150$ & $67.9$ & $7743$ & $2.27\times$\\
                 HALP-$42\%$~\cite{shen_2021} & $171$ & $\mathbf{68.3}$ & $7940$ & $2.32\times$ \\
                 \textbf{SMCP-$40\%$ (Ours)} & $208$ & $\mathbf{68.3}$ & $\mathbf{8163}$ & $\mathbf{2.39\times}$ \\
                \midrule
                \bottomrule
            \end{tabular}
        }
    \end{minipage}
\end{table}

Our method performs comparably to prior works at low pruning ratios and outperforms them for large pruning ratios. For the PyTorch ResNet50 baseline model, we achieve a \( 0.3\% \) higher Top-1 accuracy with a higher FPS at 2G and 1G FLOPs with an additional \( 0.04 \times \) and \( 0.19 \times \) speedup respectively. For the EagleEye~\cite{li_2020} baseline, our method produces models near 1G FLOPs that have a \( 0.6\% \) higher Top-1 accuracy for nearly the same FPS or a similar Top-1 accuracy while being \( 19\% \) (or \( 0.5 \times\)) faster. The results are similar for ResNet101, which is based on the PyTorch model hub baseline model. At 2G FLOPs, we get a \( 0.3\% \) higher Top-1 accuracy and an additional \( 0.03 \times \) speedup. On the already compact MobileNet-V1 model, where the desired pruning ratios are smaller, our method performs comparably to prior works; at the highest pruning ratio, we show a minor FPS improvement of \( 0.07 \times \) despite a higher FLOPs count, demonstrating the ability of the optimization problem in~\cref{sec:cost_constrained_pruning} to choose cost-constraint aware masks.

\begin{figure}[t]
  \includegraphics[width=0.49\columnwidth]{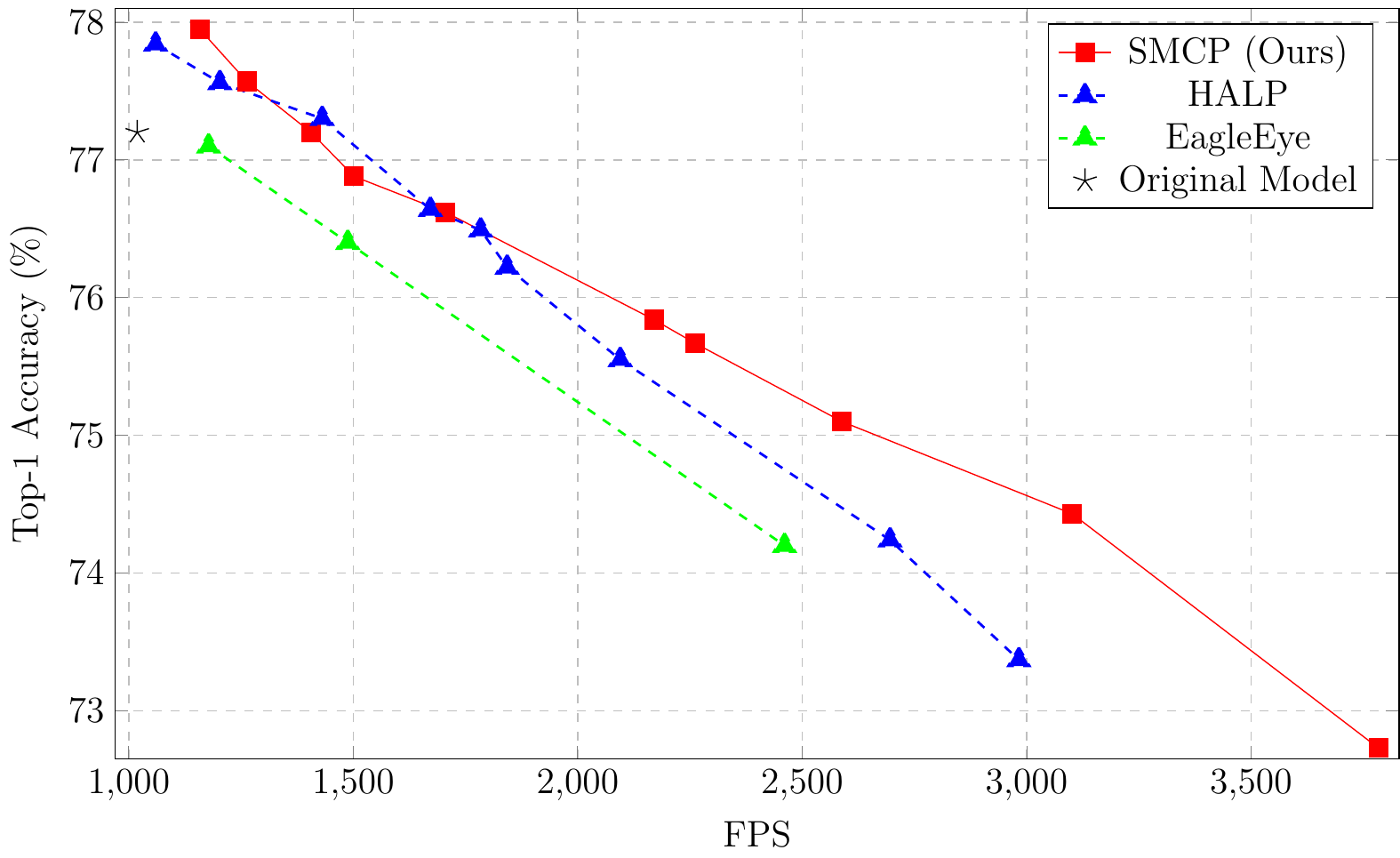}
  \includegraphics[width=0.47\columnwidth]{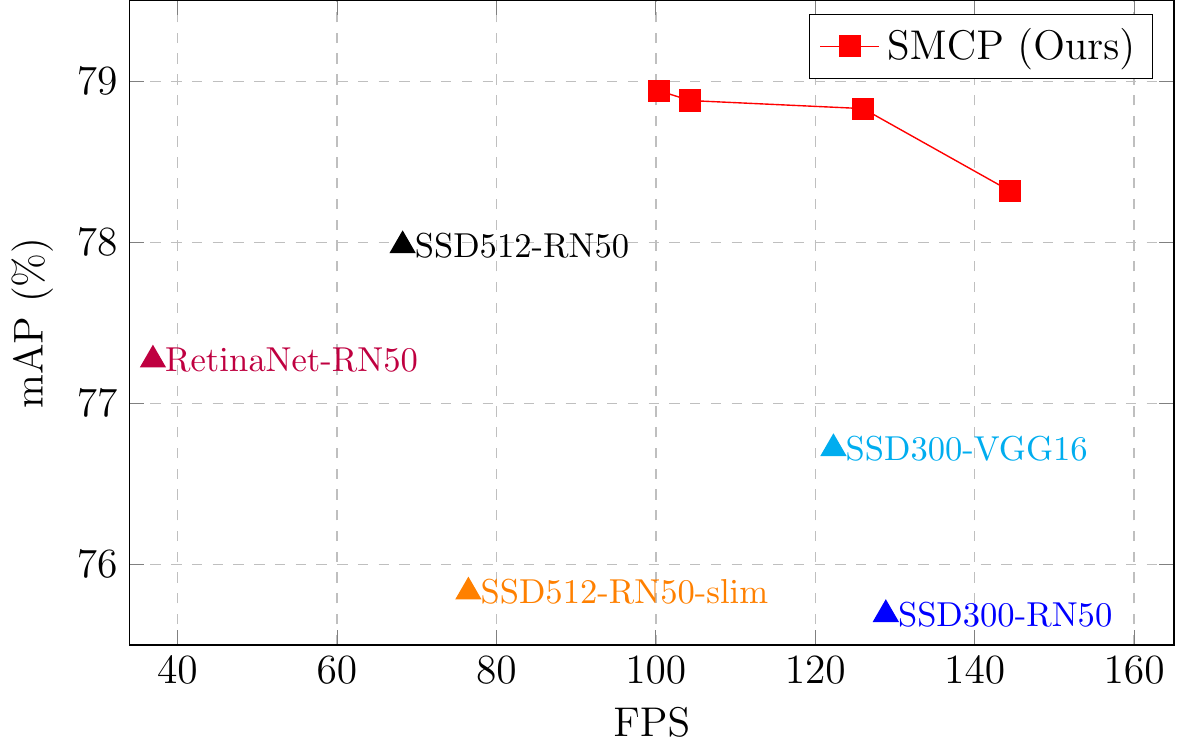}
  \caption{(Left) Top-1 accuracy tradeoff curve for pruning ResNet50 on the ImageNet classification dataset using a latency cost constraint. Baseline model is from EagleEye~\cite{li_2020}. (Right) mAP accuracy tradeoff curve for pruning SSD512-RN50 on the PASCAL VOC object detection dataset using a latency cost constraint. Top-right is better.}
  \label{fig:resnet50ee_pascal_compare}
\end{figure}

The benefits of our method, particularly at high pruning ratios, are possibly more easily seen when plotting the tradeoff curve for Top-1 accuracy versus FPS, as shown in~\cref{fig:resnet50_baseline_compare} for the PyTorch baseline and~\cref{fig:resnet50ee_pascal_compare} for the EagleEye baseline. For example in~\cref{fig:resnet50ee_pascal_compare}, at the \( 75\%\) latency reduction level (or \( 3102 \) FPS), our method outperforms the nearest HALP~\cite{shen_2021} model with a \( 0.2\% \) higher Top-1 accuracy and a \( 15 \% \) higher FPS; compared to EagleEye~\cite{li_2020}, we show a \( 0.23\% \) higher Top-1 accuracy and a \( 26 \% \) higher FPS.

Moreover, our method can aggressively prune large, over-parameterized models to outperform smaller unpruned models. As shown in~\cref{tab:latency-aware-result-compare} and~\cref{fig:resnet50ee_pascal_compare}, a \( 50\% \) pruned ResNet101 achieves a \( 1.6\% \) Top-1 improvement over a baseline ResNet50, with no performance loss, and a \( 80\% \) pruned ResNet50 achieves a similar Top-1 to an unpruned MobileNet-V1 while achieving a \( 10\% \) FPS speedup.

Lastly, the accuracy and performance gains are in part due to the final network architecture chosen by our method. In particular, since we solve a global resource allocation problem during training, our method automatically determines the layer-wise pruning ratios for the given cost function and constraint. For example, on ResNet50, we find that SMCP is aggressive in pruning the early convolution layers and leaves the later layers better preserved; we provide additional analysis and figures in the supplementary materials.

\subsection{PASCAL VOC results}
To analyze our method beyond image classification, we also analyze SMCP on the PASCAL VOC object detection dataset~\cite{everingham_2010}. Specifically, we consider whether a large model, such as SSD512~\cite{liu_2016_ssd} with a ResNet50 backbone, can be pruned at a high ratio to match the FPS of smaller models while retaining a superior mAP (mean average precision). We use the ``07+12'' train and test setup of~\cite{liu_2016_ssd} and prune both the backbone and feature layers.

As shown in~\cref{fig:resnet50ee_pascal_compare}, our method can prune an SSD512-RN50 to have a higher mAP than the pretrained model and a faster FPS than the much smaller SSD300-RN50 model, again showing the ability of our method to aggressively prune large over-parameterized models to outperform smaller models. In particular, our fastest pruned model has a \( 2.63 \) point higher mAP score while achieving \( 12\% \) higher FPS. Critically, the latency reduction to achieve this is \( 75 \% \), demonstrating the strength of our approach in the high pruning ratio regime. We also compare to and outperform a number of other common detector models. 

\subsection{Ablation study}\label{sec:ablation_study}
\begin{figure}[t]
  \begin{center}
    \includegraphics[width=0.6\columnwidth]{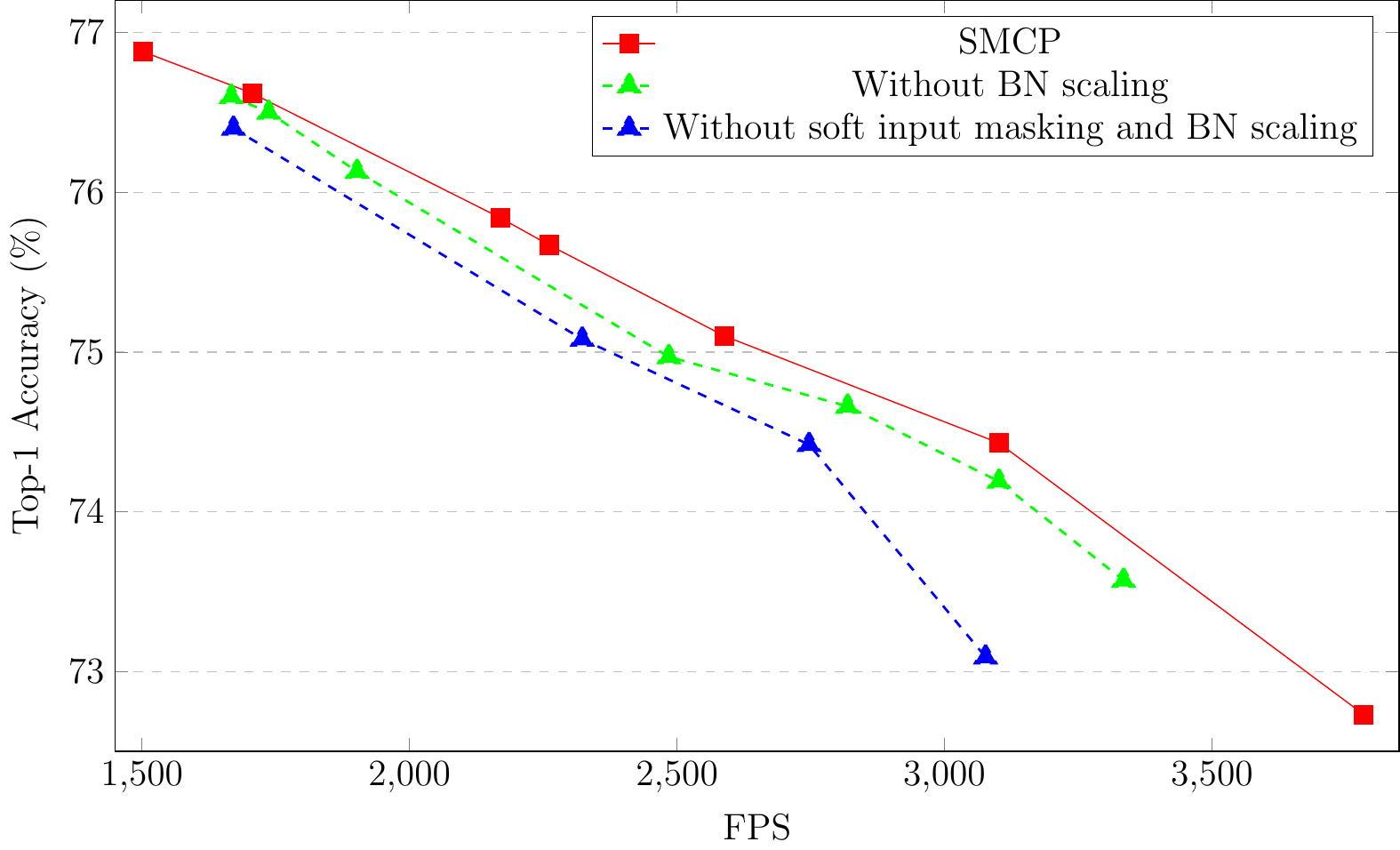}
  \end{center}
  \caption{Ablation study for SMCP at high pruning ratios on ResNet50 using the EagleEye~\cite{li_2020} baseline. We remove consecutively two major components of our method, soft input masking and batch normalization scaling, and observed worse Top-1 accuracy and FPS than the full SMCP method.}
  \label{fig:ablation_study}
\end{figure}

We also study the effect of our contributions on the accuracy results shown above, specifically at high pruning ratios. We run our method again on the ImageNet classification dataset, starting from the ResNet50 EagleEye~\cite{li_2020} baseline. We first remove the batch normalization scaling technique from~\cref{sec:bn_scaling} while keeping the soft input channel masking re-parameterization of~\cref{sec:soft_mask}. We then additionally remove the soft input channel masking, reverting to permanent pruning. We keep the solver and latency constraint in~\cref{sec:cost_constrained_pruning} unchanged. The ablation results are shown in~\cref{fig:ablation_study}.
Removing the batch normalization scaling generally leads to marginally worse results, due to the training instability described in~\cref{sec:bn_scaling}. Additionally removing the soft input masking, thereby using permanent channel pruning, degrades accuracy and performance further.

\subsection{Choice of latency cost constraint}
Although our cost-constrained formulation is general to any number of cost functions, the benefits of our approach are most pronounced under challenging, non-linear latency cost landscapes (i.e., latency cliffs for GPUs). Linear constraints (i.e., parameter/FLOP constraints) lessen the need for soft masking and an efficient and global resource allocation: removed channels are more likely to stay pruned once removed and the number of remaining channels in each layer tends to change slowly. Despite training against a latency constraint, \cref{tab:latency-aware-result-compare} shows that SMCP is comparable to or even outperforms previous methods under low FLOP constraints.



\section{Conclusion}
By applying channel pruning, modern CNNs can be significantly accelerated, with a smaller memory footprint, computational cost, and inference time. In this work, we presented a novel structured input channel pruning approach, called SMCP, that combines soft masking of input channels, a batch normalization scaling technique, and the solution to a resource allocation problem to outperform prior works. We motivate the use of each component of our method and demonstrate their effectiveness on both the ImageNet and PASCAL VOC datasets. Although we only consider channel pruning in this work, our approach can be extended to jointly consider both channel and N:M structured pruning~\cite{mishra_2021_ampere} to satisfy an explicit cost-constraint. This can be viewed as an extension of both this work and that of~\cite{zhou_2021} and is left for a future work.

%
%
\clearpage
\bibliographystyle{splncs04}
\bibliography{egbib}

\clearpage
\section{Supplementary Materials}

\subsection{Experimental settings}
We define our networks and pruning operations using PyTorch~\cite{paszke_2019} and run our experiments across \( 8 \) NVIDIA Tesla V100 GPUs using automatic mixed precision and DDP (distributed data parallel) training. 

\subsubsection{ImageNet experiments}
For all of the ImageNet experiments, across ResNet50, ResNet101, and MobileNet-v1, we follow NVIDIA's ResNet50 training setup~\cite{nvidia_resnet_training_guide} with a batch size of \( 256 \) per GPU, a linear learning rate warmup period of \( 8 \) epochs, a cosine decay learning rate schedule~\cite{loshchilov_2017}, and a \( 90 \) epoch training time.

We start pruning after \( K_w = 10 \) epochs, reach the target \( K_t = 30 \) epochs later at epoch \( 40 \), and allow the masks to continue to refine until fixing the masks for the final \( K_c = 45 \) epochs. We recompute the masks every \( r = 80 \) steps. (We conduct a small sensitivity study on these hyperparameters later in the supplementary materials). To ensure a GPU-friendly setting of the masks, we set \( \mathcal{P}^{(l)} = \{ i: i\%8=0, i \leq C^{(l)}_{in} \} \) for every layer but the first two. We prohibit any pruning of the first convolution by setting \( \mathcal{P}^{(1)} = \{ 3 \}, \mathcal{P}^{(2)} = \{ C^{(2)}_{in} \} \) and prohibit any layer pruning by ensuring \( 0 \not\in \mathcal{P}^{(l)} \).  

We build the lookup table for ResNet50 and ResNet101 with a batch size of \( 256 \) and MobileNet-V1 with a batch size of \( 512 \); we assess the latency of the final pruned networks under these settings as well.

\subsubsection{PASCAL VOC experiments}
We use the SSD512 model described in~\cite{liu_2016_ssd}, swapping the VGG16 backbone for a ResNet50 backbone. As in~\cite{huang_2017}, we keep only the first three stages of the convolutions and change the strides in the third to stage to \( 1 \times 1 \). We then add \( 6 \) pairs of feature detection layers as in~\cite{liu_2016_ssd} and the localization and confidence heads to generate the boxes and their scores. We train for \( 800 \) epochs with a batch size of \( 16 \) per GPU, using PyTorch's \textit{SyncBatchNorm} to synchronize the batch normalization statistics. We use a learning rate of \( 8\text{e-}3 \) for the total batch size \( 128 \), a linear warmup to that rate over \( 50 \) epochs, and reduce the rate by a multiple of \( 3/8, 1/3, 2/5, 1/10 \) at \( 600, 700, 740, 770 \) epochs respectively. For network biases, we double the learning rate. We also set the weight decay to \( 2\text{e-}3 \) except for the batch normalization parameters. We use the SGD optimizer with a momentum of \( 0.9 \)

We start pruning after \( K_w = 60 \) epochs, reach the target \( K_t = 250 \) epochs later at epoch \( 310 \), and allow the masks to continue to refine until fixing the masks for the final \( K_c = 350 \) epochs. We recompute the masks every \( r = 100 \) steps. We use the same rules for \( \mathcal{P}^{(l)} \) as with the ImageNet experiments.

We assess the latency of the final pruned SSD512-ResNet50 at a batch size of \( 1 \) for comparison with other detectors.

\subsection{ResNet50 layer-wise pruning ratios}
\begin{figure}[t]
  \begin{center}
    \includegraphics[width=0.9\columnwidth]{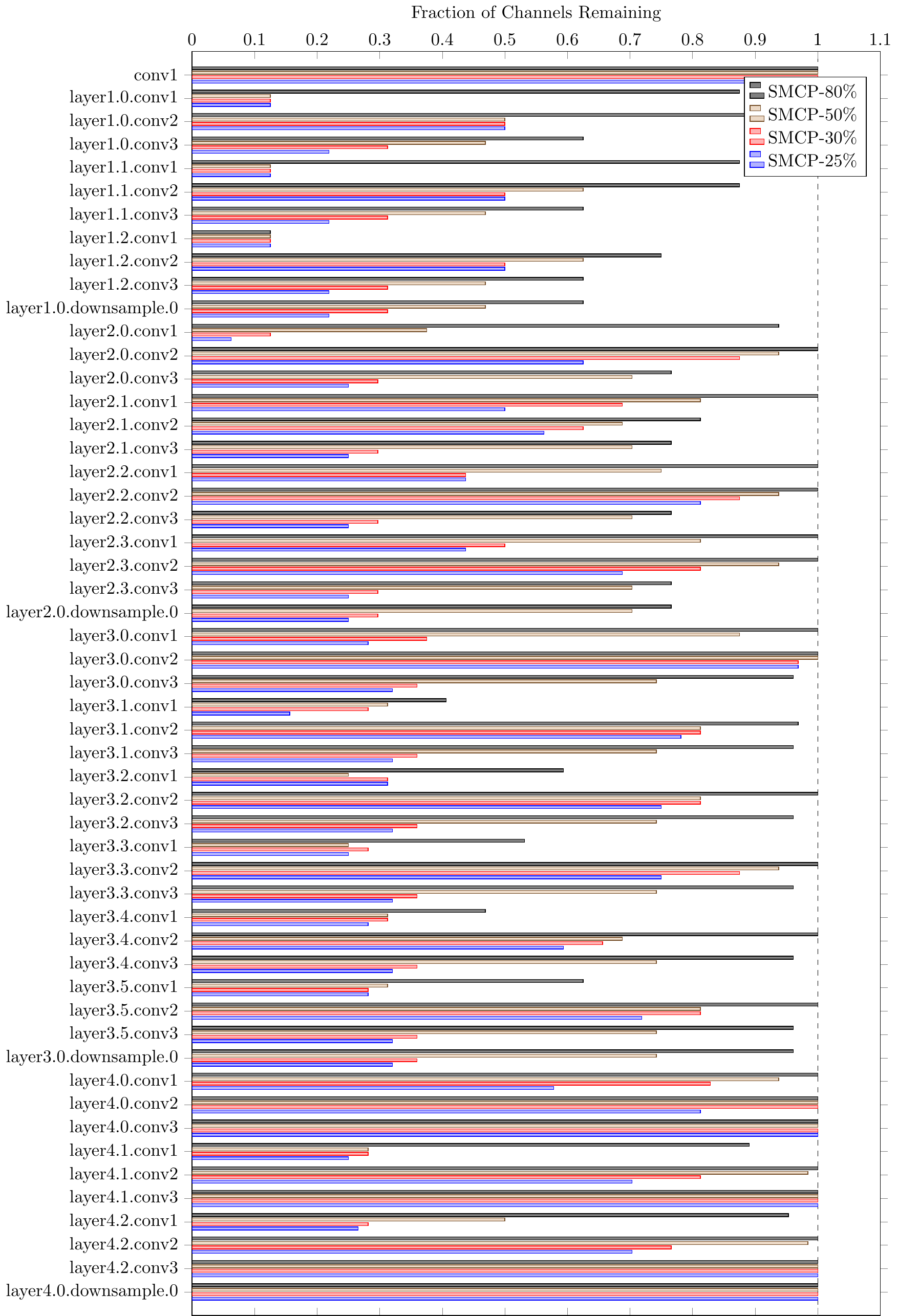}
  \end{center}
  \caption{Fraction of remaining channels per layer for SMCP models on the ImageNet classification dataset.}
  \label{fig:compression_ratios}
\end{figure}

Since we solve a global resource allocation problem, there are no preset layer-wise pruning ratios. We therefore analyze the final pruning ratios for each layer to derive insights into our method. \cref{fig:compression_ratios} plots the fraction of channels remaining in each layer relative to the original unpruned model. Generally, we find that SMCP prunes heavily in the early layers of the network and preserves more channels in the later layers of the network and in the second convolution layer in each residual block throughout (i.e., the conv2 layers).

\begin{figure}[t]
  \begin{center}
    \includegraphics[width=0.9\columnwidth]{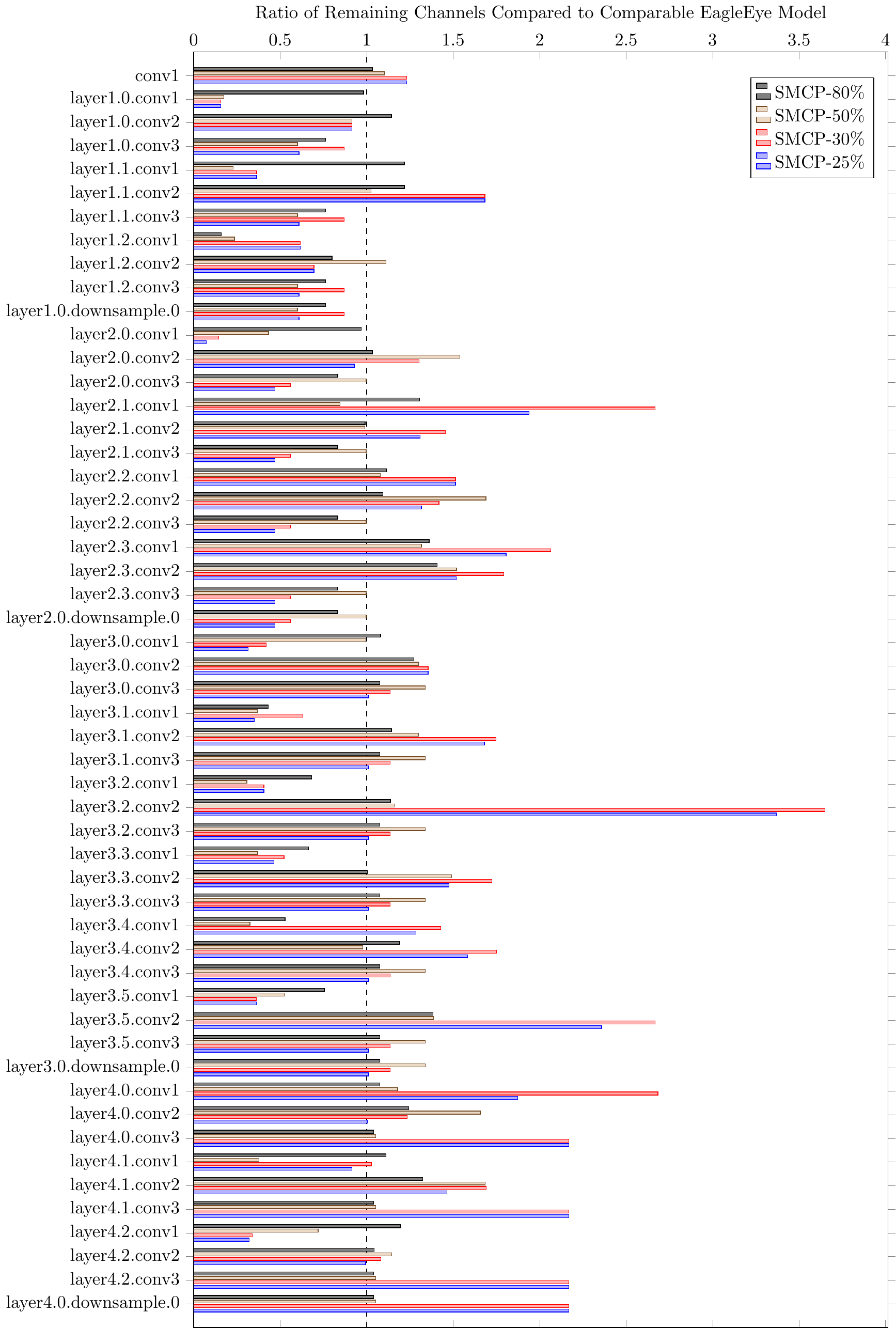}
  \end{center}
  \caption{Relative comparison of number of remaining channels per layer for SMCP and EagleEye ResNet50 models on the ImageNet classification dataset. Each SMCP model is compared to the EagleEye model of comparable FLOPs.}
  \label{fig:compression_compare}
\end{figure}

We also compare our pruning ratios to the EagleEye~\cite{li_2020} models of comparable FLOPs in \cref{fig:compression_compare}. The general pattern of pruning heavily early and lighter later still holds. In particular, at the high pruning ratios, SMCP is able to keep a much larger number of channels in the later layers of the network, often twice as many as the EagleEye model, which seems to convey a Top-1 accuracy improvement as shown in \cref{tab:latency-aware-result-compare}.

\subsection{Pruning schedule hyperparameters}
\begin{figure}[t]
  \begin{center}
    \includegraphics[width=0.8\columnwidth]{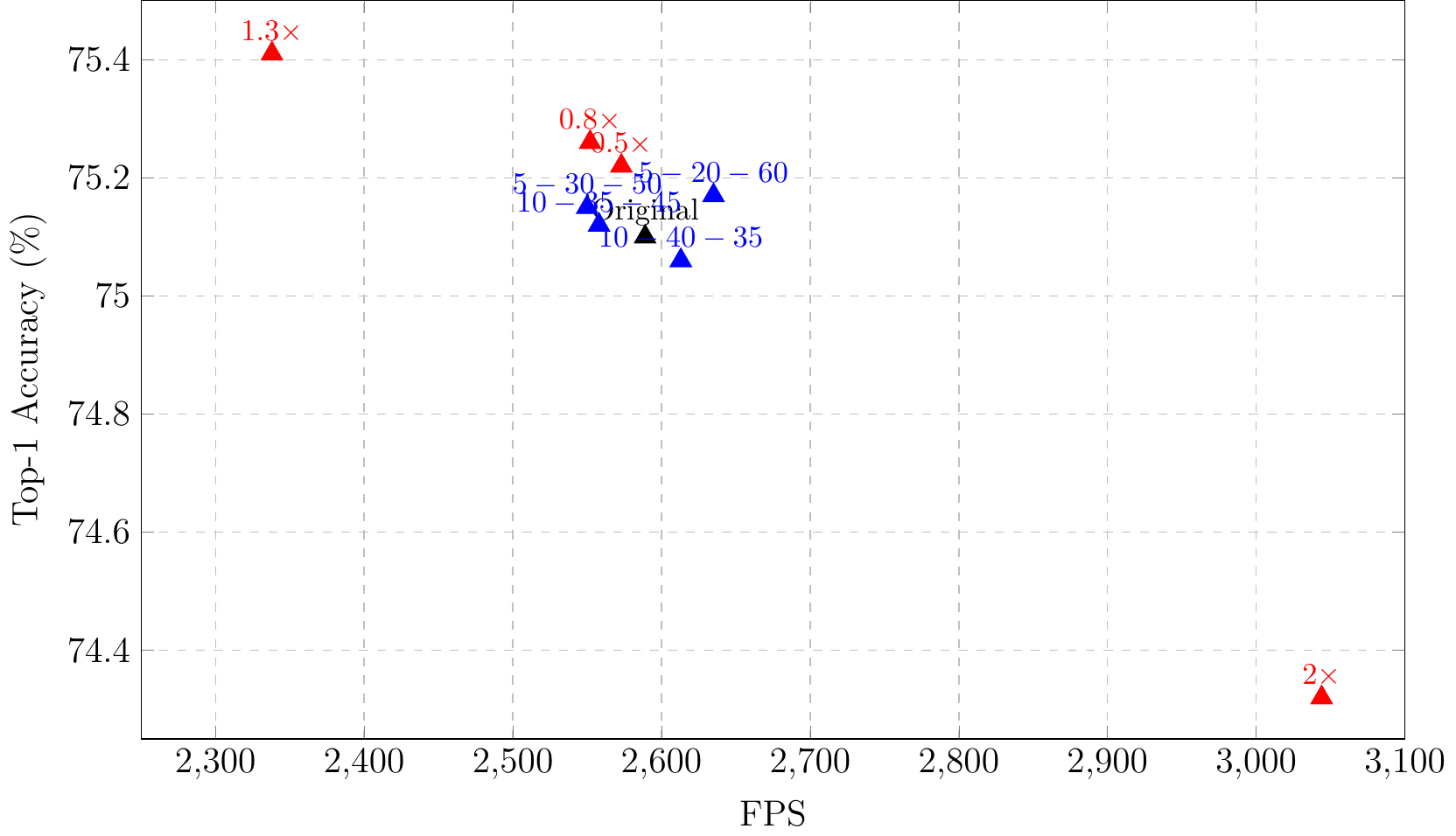}
  \end{center}
  \caption{Ablation study for SMCP's pruning schedule hyperparameters. Baseline model is EagleEye's ResNet50 model~\cite{li_2020}. Multiple denotes changing the rewiring frequency by the given multiple. Triples denote changing the target schedule hyperparameters. FPS measured on an NVIDIA TITAN V GPU.}\label{fig:rewire_ablation} 
\end{figure}

Our algorithm only introduces new additional hyperparameters for the pruning schedule: the number of steps between recomputing the masks $ r $ and the target schedule defined by $ (K_w, K_t, K_c) $. To determine if our results are sensitive to these choices, we train a ResNet50 model at a $ 70\% $ reduction while varying $ r $ for a fixed schedule and varying the schedule for fixed $ r $; the results are shown in \cref{fig:rewire_ablation}. We find that varying the target schedule, from the original schedule of $ (10, 30, 45)$ to $ (10, 35, 45), (10, 40, 35), (5, 20, 60), (5, 30, 50) $, has little impact on the final accuracy and FPS, with Top-1 by at most $ 0.07\% $ with a FPS difference at most $ 46 $ FPS. Changing $ r $ has a much bigger impact. To understand why, recall that in order to define the knapsack-like optimization problem in \cref{eqn:pruning_knapack_problem} we had to make an approximation $ \mathcal{T}^{(l)} \left( p^{(l)}, p^{(l-1)} \right) \approx \mathcal{T}^{(l)} \left( p^{(l)}, \overline{p^{(l-1)}} \right) $. Therefore solving the optimization problem less often results in a worse approximation of the loss landscape and a final model cost that can vary more from the desired cost. Nonetheless the resulting pruned model is quite accurate for its cost and lives on a Pareto frontier of possible models of different costs.

\subsection{Equivalence of Taylor FO importance scores}
\begin{theorem}\label{thm:taylorFO_input}
    For a network composed of Conv-BN-ReLU blocks and without a mask re-parameterization, the first-order Taylor importance on the batch normalization weight and bias (Taylor-FO-BN~\cite{molchanov_2019}) of layer \( l \) is equivalent to a first-order Taylor importance on the weights of the downstream input channel in layer \( l + 1 \).
    \begin{equation}
        \gamma^{(l)}_i g_{\gamma^{(l)}_i} + \beta^{(l)}_i g_{\beta^{(l)}_i} = \sum_{o, r, s} W^{(l+1)}_{o,i,r,s} g_{W^{(l+1)}_{o,i,r,s}}
    \end{equation}

    \begin{proof}
        Let the output BN layer be parameterized by \( \gamma \) and \( \beta \) and the following convolution layer be parameterized by \( W \). Suppose \( y \) is the input to the BN layer, \( z \) is the output of the BN layer and \( x = ReLU(y) \) is the input to the convolution layer. From the definition of a batch normalization layer, it follows that the gradient of the batch normalization weight and bias are given by
        \begin{align}
            g_{\gamma_i} & = \sum_{m,n} g_{z_{i,m,n}} \hat{y}_{i,m,n} \\
            g_{\beta_i} & = \sum_{m,n} g_{z_{i,m,n}}
        \end{align}
        where \( \hat{y} \) is the value of \( y \) after normalization.
        Then,
        \begin{align}
            \gamma_i g_{\gamma_i} + \beta_i g_{\beta_i} &= \sum_{m,n} \left( \gamma_i \hat{y}_{i,m,n} + \beta_i \right) g_{z_{i,m,n}} \\
            &= \sum_{m,n} z_{i,m,n} g_{z_{i,m,n}} \\
            &= \sum_{m,n} x_{i,m,n} g_{x_{i,m,n}} \\
            &= \sum_{o,r,s} W_{o,i,r,s} g_{W_{o,i,r,s}}
        \end{align}
        where the last step proceeds from the definition of the convolution.
    \end{proof}
\end{theorem}
\begin{corollary}
    This holds even in the presence of skip connections in architectures like ResNet.
    \begin{proof}
        Suppose for example that there are \( k \) BN layers whose outputs \( z^{(k)} \) are added to create \( z \), applied through a ReLU nonlinearity to create \( x \), and then distributed as \( x^{(j)}=x \) to \( j \) different convolution layers with weights \( W^{(j)} \). Using results from the proof of~\cref{thm:taylorFO_input}, we have
        \begin{align}
            \sum_k \gamma^{(k)}_i g_{\gamma^{(k)}_i} + \beta^{(k)}_i g_{\beta^{(k)}_i} & = \sum_{k,m,n} z^{(k)}_{i,m,n} g_{z^{(k)}_{i,m,n}} \\
            & = \sum_{k,m,n} z^{(k)}_{i,m,n} g_{z_{i,m,n}} \\
            & = \sum_{m,n} z_{i,m,n} g_{z_{i,m,n}} \\
            &= \sum_{m,n} x_{i,m,n} g_{x_{i,m,n}} \\
            &= \sum_{j,m,n} x^{(j)}_{i,m,n} g_{x^{(j)}_{i,m,n}} \\
            &= \sum_{j,o,r,s} W^{(j)}_{o,i,r,s} g_{W^{(j)}_{o,i,r,s}}
        \end{align}
        where we use gradient rules for both the addition and ReLU operations to simplify.
    \end{proof}
\end{corollary}
\begin{corollary}
    \cref{thm:taylorFO_input} also holds under hard masking (masking without the STE) and for unpruned channels under soft masking. For pruned channels under soft masking, \( \gamma_i g_{\gamma_i} + \beta_i g_{\beta_i} = 0 \) since it is the sparse weights that determine the gradient of input feature map but the dense weights receive a dense gradient update.
\end{corollary}

\subsection{Exploding gradients without batch normalization scaling}
Let's consider the popular Conv-BN pattern, with weights \( W, \gamma, \beta \) and statistics \( \mu^{(t)}, \sigma^{(t)} \) at step \( t \). Suppose that a fraction \( 1 - \alpha \) of the input channels are pruned at the end of one training step. During the next training step, there are fewer non-zero weights, which can intuitively cause the batch variance to shrink: \( \sigma^{(t+1)} < \sigma^{(t)} \). This does not affect the forward pass significantly, in that the output of the BN layer is still roughly \( \mathcal{N}\left( \beta, \gamma^2 \right) \), but it affects the gradients quite noticeably. In fact, this causes the gradients flowing to the remaining, unpruned channels to be boosted quite significantly as \( \alpha \) nears \( 1 \).

Concretely, let the Conv-BN pair be characterized by \( z = W * x, \hat{z} = \frac{z^ - \mu}{\sigma}, y = \gamma \hat{z} + \beta \). The gradient to the intermediate feature map \( z \) is equal to
\begin{equation}
    g_{z_{o,h,w}} = \frac{\gamma_o}{\sigma_o} \left( g_{y_{o,h,w}} - \frac{1}{S} g_{\beta_o} - \frac{1}{S} \hat{z}_{o,h,w} g_{\gamma_o} \right)
\end{equation}
where \( S \) is the size of each channel in the feature map \( y \). When \( \sigma_o \) shrinks, the gradient magnitude is inversely increased, causing \( g_{z_{o,h,w}} \) at time \( t + 1 \) to be roughly a factor of \( s_t = \sigma^{(t)} / \sigma^{(t+1)} \) larger than it was at time \( t \). In the extreme case with \( \alpha \rightarrow 1 \), we can get exploding gradients, with a degeneracy at \( \alpha = 1 \) which corresponds to layer pruning.

\subsection{Derivation of the cost-constrained optimization problem}\label{sec:opt_derivation}
We start from the idea cost-constrained network objective for neural network \( f: X \rightarrow Y \)
\begin{equation}
    \argmin_{W} \mathcal{L}\left(W, \mathcal{D} \right) \text{ s.t. } \mathcal{T}\left(f(W, x_i)\right) \leq \tau
\end{equation}
where \( \mathcal{L} \) is the network loss function, \( W = \{ W^{(l)} \} \) are the network's weights, \( \mathcal{D} = \{ (x_i, y_i) \} \) is the training set, \( \mathcal{T} \) is the network's cost function, and \( \tau \) is the cost constraint. Using the input channel masks \( M = \{ m^{(l)} \} \) and sparse weights \( \widetilde{W}^{(l)} = W^{(l)} \odot m^{(l)} \), as described in~\cref{sec:soft_mask}, we get a joint optimization problem over the weights and masks
\begin{equation}
    \argmin_{W, M} \mathcal{L}\left(\widetilde{W}, \mathcal{D} \right) \text{ s.t. } \mathcal{T}(M) \leq \tau
\end{equation}
where the cost constraint is now only a function of the masks. This is a combinatorially hard discrete optimization problem over the masks, so to make it tractable, we make several changes. First, we replace the loss minimization objective with an importance maximization objective and linearize it with a per-channel importance score \( \mathcal{I}^{(l)}_i \) defined in~\cref{eqn:taylorfo_channel}. This assumes, that despite the nonlinearities of \( f \), the importance score \( \mathcal{I}^{(l)}_i \) is a good approximation of the effect of removing that channel from the network. This makes the objective linear in the masking variables:
\begin{equation}
    \argmax_{M} \sum_{l=1}^L \sum_{i=1}^{C^{(l)}_{in}} \mathcal{I}^{(l)}_i m^{(l)}_i \text{ s.t. } \mathcal{T}(M) \leq \tau
\end{equation}
Second, we assume the cost function \( \mathcal{T} \) is layer-wise separable into constituent cost functions \( \mathcal{T}^{(l)} \) that depend only on the number of input and output channel masks for that layer. The output channel mask is defined by the input channel mask of the downstream layer. This yields
\begin{align}
    \argmax_{M} & \sum_{l=1}^L \sum_{i=1}^{C^{(l)}_{in}} \mathcal{I}^{(l)}_i m^{(l)}_i \\
    \textrm{s.t.} \quad & \sum_{l=1}^L \mathcal{T}^{(l)}\left( \norm{m^{(l)}}_1, \norm{m^{(l+1)})}_1 \right) \leq \tau \nonumber
\end{align}
Lastly, we add the additional constraint on the allowable values for the number of input channels, \( \norm{m^{(l)}}_1 \in \mathcal{P}^{(l)} \), to get~\cref{eqn:original_opt_problem}.

\subsubsection{Skip connections}\label{sec:skip-connections}
For architectures that have skip connections or other structural branching features, the optimization problem in~\cref{eqn:original_opt_problem} needs one additional constraint. Specifically, all layers that share the same input channels must prune identically to one another. For example, in a ResNet bottleneck block that performs downsampling, both the downsample convolution and the first convolution in the branch share the same input channels. Therefore, their masks \( m^{(down)} \) and \( m^{(conv1)} \) must be equal to each. More formally, let \( g_k \) be a group of layers that share input channels. Then, we must add the constraint
\begin{equation}
    m^{(l)} = m^{(g_k)} \quad \forall l \in g_k
\end{equation}
for every group \( g_k \) of layers in the network. For \( G = \{ g_k \} \), the optimization problem reduces to choice of masks over each group instead of each layer
\begin{align}
        \argmax_{M} & \sum_{k=1}^{|G|} \sum_{i=1}^{C^{(g_k)}_{in}} \left( \sum_{l \in g_k}  \mathcal{I}^{(l)}_i \right) m^{(g_k)}_i \\
        \textrm{s.t.} \quad & \sum_{k=1}^{|G|} \sum_{l \in g_k} \mathcal{T}^{(l)}\left( \norm{m^{(g_k)}}_1, \norm{\overline{m^{(l+1)})}}_1 \right) \leq \tau \nonumber \\
        & \norm{m^{(g_k)}}_1 \in \bigcup_{l \in g_k} \mathcal{P}^{(l)} \nonumber \\
        & m^{(l)} = m^{(g_k)} \quad \forall l \in g_k \nonumber
\end{align}
where we decouple the cost impacts of masks in consecutive layers as in~\cref{eqn:pruning_knapack_problem}. This is a simply a generalization of~\cref{eqn:original_opt_problem} where we consider groups of layers together instead of each layer individually.

\subsection{Cost-constrained channel pruning as multiple-choice knapsack}
Our cost-constrained resource allocation problem in~\cref{eqn:pruning_knapack_problem} is an example of a general class called the multiple-choice knapsack problem of~\cite{sinha_1979}. We now show this connection explicitly. We start with~\cref{eqn:pruning_knapack_problem}, reproduced here for readability,
\begin{align}
    \max_{p^{(2)}, \dots, p^{(L)}} \quad & \sum_{l=1}^L \sum_{i=1}^{p^{(l)}} \mathcal{I}^{(l)}_{(i)} \nonumber \\
    \textrm{s.t.} \quad & \sum_{l=1}^L \mathcal{T}^{(l)}\left( p^{(l)}, \overline{p^{(l+1)}} \right) \leq \tau \nonumber \\
    & p^{(l)} \in \mathcal{P}^{(l)}. \nonumber
\end{align}
Now, we define
\begin{align}
    v_{l, j} & = \sum_{i=1}^{j} \mathcal{I}^{(l)}_{(i)} \\
    c_{l, j} &= \mathcal{T}^{(l)}(j, \overline{p^{(l+1)}}) \\
    x_{l, j} &= \mathbf{1}\left( j = p^{(l)} \right)
\end{align}
where \( \mathbf{1}(\cdot) \) is the indicator function.
This yields the the equivalent optimization problem
\begin{align}
    \max_{x} \quad & \sum_{l=1}^L \sum_{j=1}^{n_l} v_{l, j} x_{l, j} \\
    \textrm{s.t.} \quad & \sum_{l=1}^L \sum_{j=1}^{n_l} c_{l, j} x_{l, j} \leq \tau \nonumber \\
    & x_{l, j} \in \{0,1\}, \quad \sum_{j=1}^{n_l} x_{l, j} = 1 \nonumber \\
    & \sum_{j \in \mathcal{P}^{(l)}} x_{l, j} = 1 \nonumber
\end{align}
where \( n_l = C^{(l)}_{in} \) is the maximum number of input channels for layer \( l \). By trimming the problem to only the values \( v_{l,j} \) and costs \( c_{l,j} \) where \( j \in \mathcal{P}^{(l)} \), we recover the general form of the multiple choice knapsack problem shown in~\cref{eqn:multple_choice_knapsack}.

\subsection{Solving the multiple-choice knapsack problem}
The standard method for solving the classic 0-1 knapsack problem is a dynamic programming algorithm. For the multiple-choice knapsack problem~\cite{sinha_1979}, Dudzi\'{n}ski and Walukiewicz~\cite{dudzinski_1987} define a similar dynamic programming solution. It requires integer costs \( c_{l,i}, C \in \mathbb{Z}_{\geq 0} \) and has solution runtime complexity \( O( n C) \) where \( n = \sum_g n_l \) and space complexity \( O(G C) \) (in order to recover the items used) when \( c_{l,i}, C \in \mathbb{Z}_{\geq 0} \). When the costs are not integer or are incredibly large integers, this becomes intractable unless a scaling and rounding step is performed. Even still, for problems with very sparse values \( v_{l,i} \) and costs \( c_{l,i} \), the dynamic programming approach is inefficient. We instead solve the MCK problem using a GPU-implemented generalization of the meet-in-the-middle algorithm used for the classic 0-1 knapsack problem. The full algorithm is defined in \cref{alg:multiple_choice_knapsack}. The benefit of this approach is we only store feasible values and costs and aggressively reject suboptimal solutions with the condense step. However, the runtime and space complexities are asymptotically much worse in general, as shown in~\cref{thm:mck_mim_complexity}.

\begin{theorem}\label{thm:mck_mim_complexity}
    The meet-in-the-middle algorithm in~\cref{alg:multiple_choice_knapsack} has worst case runtime complexity \( O \left( L B^L \log(B) \right) \) and space complexity \( O \left( B^L \right) \) where \( B = \max_l n_l \) and we assume \( B \gg L \).

    \begin{proof}
        We start by deriving the complexity of the merge function for a group of size \( M \) and another of size \( N \).
        \begin{align}
            T_{\textrm{merge}}(M, N) & = O(M N) + T_{\textrm{condense}}(M N) \\
            & = O( M N \log(M N))
        \end{align}
        since the condense function requires and is dominated by a sort of the values.
        
        The runtime complexity of the full multiple-choice knapsack solver can be bounded by assuming every group takes the maximum size \( B \). Without loss of generality, we also assume \( L = 2^j \) for some \( j \). Then, for a multiple-choice knapsack (MCK) problem with \( L \) groups of size \( B \), we have
        \begin{alignat}{3}
            T_{\textrm{mck}}(L, B) & = && T_{\textrm{mck}}\left( \frac{L}{2}, B^2 \right) + \frac{L}{2} T_{\textrm{merge}}(B, B) \\
            & \leq && T_{\textrm{mck}}\left( 1, B^{2^j} \right) \\
            & && + \sum_{i=1}^j \frac{L}{2^i} T_{\textrm{merge}}\left( B^{2^{i-1}}, B^{2^{i-1}} \right) & \\
            & \leq && O\left( B^{2^j} \right) \\
            & && + \sum_{i=1}^j \frac{L}{2^i} 2^i O\left( B^{2^i} \log(B) \right) \\
            & \leq && O\left( L B^L \log(B) \right)
        \end{alignat}
        since the merge can create a new item for every combination of items in the two merging groups (squaring the size of the group), \( T_{mck}(1, B) = O(B) \), and the runtime is dominated by the final merge of groups.
        
        The space complexity proceeds similarly, except for the \( \log \) factor which is due to the sort in the condense step.
    \end{proof}
\end{theorem}
\begin{corollary}\label{thm:mck_mim_complexity_improved}
    The runtime and space complexity of \cref{alg:multiple_choice_knapsack} can be improved to \( O \left( L B^{L/2} \log(B) \right) \) and \( O \left( B^{L/2} \right) \) respectively by replacing the final merge with a \( O \left( B^{L/2} \log\left( B^{L/2} \right) \right) \) sort and sweep over the last two groups.
\end{corollary}

\begin{algorithm}
    \caption{MCK meet-in-the-middle solver}\label{alg:multiple_choice_knapsack}
    \textbf{Input}: Number of groups \( L \), group value vectors \( v_{l} \), group cost vectors \( c_{l} \leq C \), and capacity \( C \) \\
    \textbf{Output}: Best value \( v_{best} \), best cost \( c_{best} \), and used items \( k_l = i \textrm{ s.t. } x_{l,i} = 1 \).
    \begin{algorithmic}
        \Function{MCK}{$v, c, C $}
            \State \( L \gets \textrm{len}(v) \)
            \State \( k_l \gets 0 \quad \forall l \in [L] \)
            \If{\( L == 1 \)}
                \State \( k_1 \gets \argmax_i v_{1,i} \)
                \State \Return \( v_{1, k_1}, c_{1, k_1}, k \)
            \EndIf
            \For{\( l \in \textrm{range}(1, \floor{L/2}) \)}
                \State \( v_l, c_l, u_l, u_{L - l + 1} \)
                \State \( \qquad \gets \textrm{MERGE}\left( v_l, c_l, v_{L - l + 1}, c_{L - l + 1}, C \right) \)
            \EndFor
            \State \( v \gets \{ v_1, \dots, v_{\ceil{L/2}} \} \), \( c \gets \{ c_1, \dots, c_{\ceil{L/2}} \} \)
            \State \( v_{best}, c_{best}, k_{rest} \gets \textrm{MCK}(v, c, C) \)
            \For{\( l \in \textrm{range}(1, \floor{L/2}) \)}
                \State \( i \gets k_l \)
                \State \( k_l \gets u_{l, i} \)
                \State \( k_{L - l + 1} \gets u_{L - l + 1, i} \)
            \EndFor
            \State \Return \( v_{best}, c_{best}, k \)
        \EndFunction
        
        \Function{merge}{$ v_1, c_1, v_2, c_2, C $}
            \State \( v_{new, N (i-1) + j} \gets v_{1, i} + v_{2, j} \quad \forall i \in [M], j \in [N] \)
            \State \( c_{new, N (i-1) + j} \gets c_{1, i} + c_{2, j} \)
            \State \( v_{new}, c_{new}, u \gets \textrm{CONDENSE}(v_{new}, c_{new}, C) \)
            \State \( u_1 \gets \floor{\frac{u}{N}} \)
            \State \( u_2 \gets u \% N \)
            \State \Return \( v_{new}, c_{new}, u_1, u_2 \)
        \EndFunction
        
        \Function{condense}{$ v, c, C $}
            \State \( u \gets \{ i : v_{i} > v_{j} \, \forall j \neq i \text{ s.t. } c_{j} \leq c_{i} \} \)
            \State \( \qquad \cup \{ i : v_{i} \geq v_{j} \, \forall j \neq i \text{ s.t. } c_{j} = c_{i} \} \)
            \State \( u \gets u \cap \{ i : c_i \leq C \} \)
            \State \Return \( \{ v_{i} : i \in u \}, \{ c_{i} : i \in u \}, u \)
        \EndFunction
    \end{algorithmic}
\end{algorithm}

Since the multiple-choice knapsack problem reverts to the original knapsack problem under \( B=2, v_{l,1} = 0, c_{l,1} = 0 \), \cref{alg:multiple_choice_knapsack} recovers the runtime and space complexities of the standard meet-in-the-middle knapsack solver, which are \( O\left( L 2^{L/2} \right) \) and \( O\left( 2^{L/2} \right) \) respectively, using the sort and sweep improvement of~\cref{thm:mck_mim_complexity_improved}.

\subsection{Multiple-choice knapsack solve effort}
We present representative timings for the solve effort of the meet-in-the-middle multiple-choice knapsack solver in~\cref{alg:multiple_choice_knapsack}. When pruning a ResNet50, there are 38 pruning groups (see~\cref{sec:skip-connections} for more details). There are a total of \( 22,531 \) input channels, including the \( 3 \) image channels of the first convolution layer. The largest layer has \( 2048 \) input channels. Restricting the possible masks to \( 8x \) for GPU tensorcores, we test the solution time for both the meet-in-the-middle solver in~\cref{alg:multiple_choice_knapsack} and an implementation of the dynamic programming (DP) approach~\cite{dudzinski_1987}. We use a representative latency capacity of \( 255.4 \), in units of milliseconds, and randomly choose the current mask settings of the network. To use the DP approach, we define a scaling factor \( s \) and convert all costs to integers according to \( \floor{c_{l,i} s} \). For \( s = 10^d \), the DP solution will be correct to \( d \) digits;~\cref{alg:multiple_choice_knapsack} is correct up to machine roundoff. The solution times are shown in~\cref{tab:mck-solve-times}.

\begin{table}[t]
    \centering
    \begin{tabular}{lccccc}
        \toprule
        Method & Accuracy & Solve time (s) \\ \midrule
        DP (\( s=10) \) & \(1\) decimal & \(<1\) second \\
        DP (\( s=100) \) & \(2\) decimal & \(\sim 6\) second \\
        DP (\( s=500) \) & \(2-3\) decimals & \(\sim 30\) second \\
        DP (\( s=1000) \) & \(3\) decimals & \(\sim 70\) second \\
        \cref{alg:multiple_choice_knapsack} & machine roundoff & \(<1\) second \\ \bottomrule
    \end{tabular}
    \caption{Representative time to solve~\cref{eqn:pruning_knapack_problem} under different settings. DP is our implementation of the algorithm in~\cite{dudzinski_1987}.}
    \label{tab:mck-solve-times}
\end{table}

\subsection{Allowing layer pruning}
In deriving our optimization problem in~\cref{eqn:original_opt_problem}, as shown in~\cref{sec:opt_derivation}, we had to assume that the importance and cost functions were layer-wise separable, meaning the input mask \( m^{(l)} \) for layer \( l \) only affects layer \( l \) and the layer(s) immediately upstream. This assumption is obviously broken when we allow layer pruning, \( m^{(l)} = 0 \), as pruning the entire layer effectively prunes all layers in that branch of the network. 

\begin{figure}[t]
  \begin{center}
    \includegraphics[width=0.6\columnwidth]{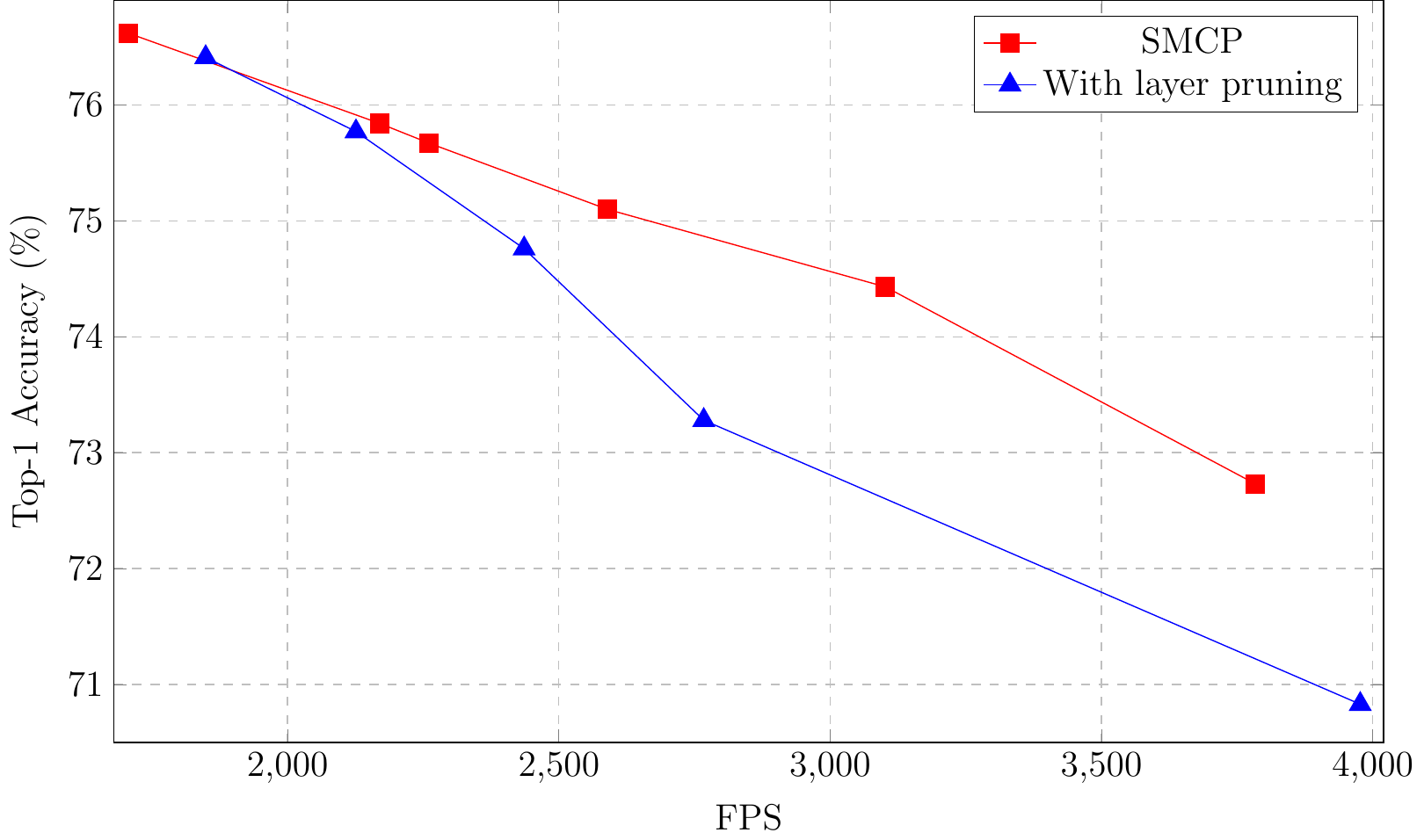}
  \end{center}
  \caption{Comparison of allowing versus disallowing layer pruning at high pruning ratios for ResNet50 on the ImageNet classification dataset using a latency cost constraint. Baseline model is from EagleEye~\cite{li_2020}. Accuracy against FPS speed shows the disadvantage of allowing layer pruning. Top-right is better. FPS measured on an NVIDIA TITAN V GPU.}
  \label{fig:layer_pruning}
\end{figure}

Nonetheless, we ran several experiments where we allowed layer pruning to occur, if so chosen by the knapsack solver, when pruning the ResNet50 from the EagleEye~\cite{li_2020} baseline. The results are shown in~\cref{fig:layer_pruning}. Breaking the theoretical layer-wise separability assumption yields poor experimental results.

\end{document}